\documentclass{article}

    \PassOptionsToPackage{numbers, sort&compress}{natbib}
\usepackage[final]{neurips_2024}

\usepackage[utf8]{inputenc} % allow utf-8 input
\usepackage[T1]{fontenc}    % use 8-bit T1 fonts
\usepackage{hyperref}       % hyperlinks
\usepackage{url}            % simple URL typesetting
\usepackage{booktabs}       % professional-quality tables
\usepackage{amsfonts}       % blackboard math symbols
\usepackage{nicefrac}       % compact symbols for 1/2, etc.
\usepackage{microtype}      % microtypography
\usepackage[table]{xcolor}         % colors

\definecolor{NavyBlue}{rgb}{0.0, 0.0, 0.5}

\usepackage{xspace}
\usepackage{multirow}
\usepackage{graphicx}
\usepackage{enumitem}
\usepackage{tabularx}
\usepackage{array}

\usepackage{amsmath,amsfonts,bm}

\def\vc{{\bm{c}}}

\def\vp{{\bm{p}}}

\def\vv{{\bm{v}}}
\def\vw{{\bm{w}}}
\def\vx{{\bm{x}}}

\def\mP{{\bm{P}}}

\newcommand{\eg}{\emph{e.g.}}
\newcommand{\ie}{\emph{i.e.}}

\graphicspath{{./}}
\newcommand{\method}{AAPE\xspace}

\title{Aggregate-and-Adapt Natural Language Prompts\\for Downstream Generalization of CLIP}

\author{%
  Chen Huang, Skyler Seto, Samira Abnar, David Grangier, Navdeep Jaitly \& Josh Susskind\\
  Apple\\
  \texttt{\{chen-huang,sseto,abnar,grangier,njaitly,jsusskind\}@apple.com}
}

\begin{document}

\maketitle

\begin{abstract}

Large pretrained vision-language models like CLIP have shown promising generalization capability, but may struggle in specialized domains (\eg,~satellite imagery) or fine-grained classification (\eg, car models) where the visual concepts are unseen or under-represented during pretraining. Prompt learning offers a parameter-efficient finetuning framework that can adapt CLIP to downstream tasks even when limited annotation data are available. In this paper, we improve prompt learning by distilling the textual knowledge from natural language prompts (either human- or LLM-generated) to provide rich priors for those under-represented concepts. We first obtain a prompt ``summary'' aligned to each input image via a learned prompt aggregator. Then we jointly train a prompt generator, optimized to produce a prompt embedding that stays close to the aggregated summary while minimizing task loss at the same time. We dub such prompt embedding as \textbf{A}ggregate-and-\textbf{A}dapted \textbf{P}rompt \textbf{E}mbedding (\textbf{\method}). \method is shown to be able to generalize to different downstream data distributions and tasks, including vision-language understanding tasks (\eg,~few-shot classification, VQA) and generation tasks (image captioning) where \method achieves competitive performance. We also show \method is particularly helpful to handle non-canonical and OOD examples. Furthermore, \method learning eliminates LLM-based inference cost as required by baselines, and scales better with data and LLM model size.
\end{abstract}

\section{Introduction}

Most existing vision-language tasks rely on large pretrained models like CLIP~\citep{radford2021learning}, which are often adapted to downstream tasks using a small amount of labeled data (as compared to the web-scale pretraining data). This is shown by many studies (\citep{zhou2021coop,zhou2022cocoop}) to be likely to obtain poor generalization performance in special domains, such as satellite imagery and fine-grained classification of car models or flower species. Such overfitting behavior is a result of limited data for those \emph{tail class concepts} in both pretraining and downstream tasks. The domain gap between pretraining and downstream data further compounds the generalization problem. For instance, CLIP may not see enough image-text pairs to identify different car models during pretraining. This makes downstream generalization to fine-grained car models difficult, especially in low-data scenarios.

\begin{figure}[!t]
\vskip -0.3in
\begin{center}
\centerline{\includegraphics[width=1.0\linewidth]{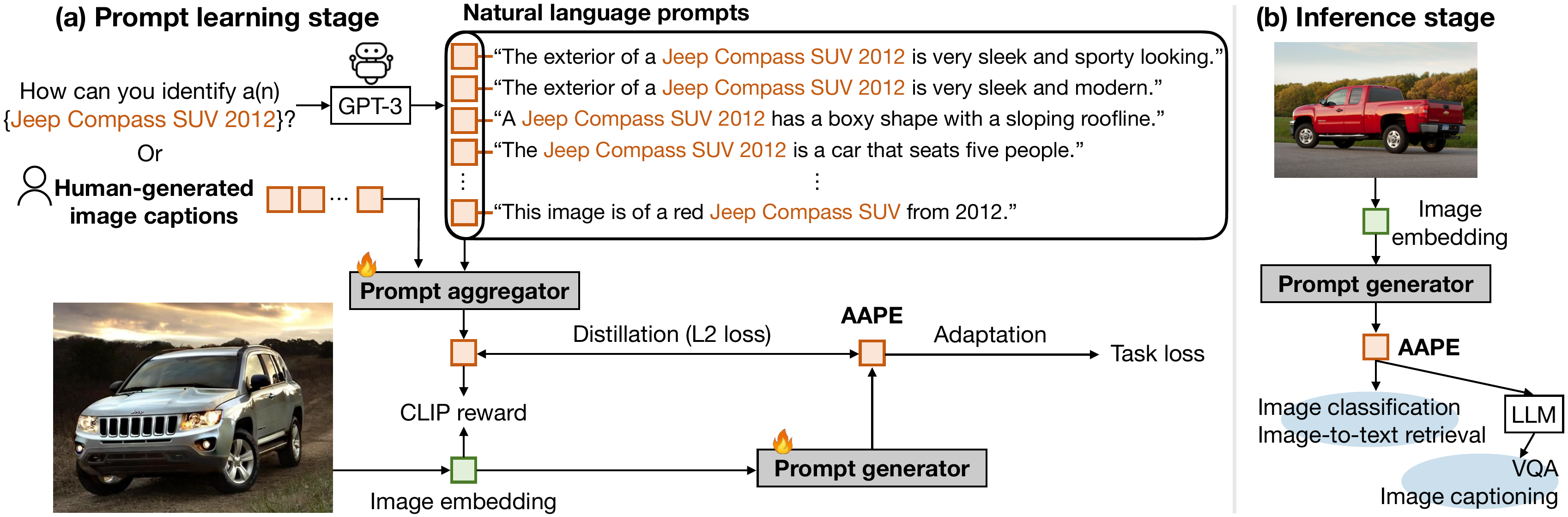}}
% \vskip -0.1in
\caption{Aggregate-and-adapt the textual knowledge in natural language prompts for downstream tasks. \textbf{(a)} For classification of object-centric images, we query GPT-3 to obtain a list of prompts for each class,~\eg,~the car model of ``Jeep Compass SUV 2012''. Note how redundant the reference prompts can be (\eg,~the first two), and how they can be irrelevant to the image (\eg,~the last prompt). Alternatively, for complex tasks like VQA, we use human-generated image captions to depict multi-object images. For all tasks, we first learn to aggregate the reference prompts into an image-aligned ``summary'' (prompt embedding) based on CLIP reward. Then a prompt generator is jointly trained to generate Aggregate-and-Adapted Prompt Embedding (\method), such that the distance between \method and the aggregated summary is minimized and the task loss is minimized too for adaptation purpose. \textbf{(b)} At test time, we only keep the prompt generator with the prompt aggregator discarded. Our \method is applicable to different vision-language tasks with strong generalization performance.}
\label{fig:schema}
\end{center}
\vskip -0.2in
\end{figure}

In this paper, we investigate using \emph{pure text-based knowledge} to boost the downstream generalization of CLIP over different data distributions and tasks, with a special focus on few-shot and OOD tasks. Note similar ideas have been explored in recent works that leverage the implicit textual knowledge in Large Language Models (LLMs) to aid vision-language tasks. For example in~\citep{yang2021empirical,su2022language}, GPT-2/3~\citep{brown2020language} is used to generate image descriptions for the tasks of Visual Question Answering (VQA) and image captioning. While for CLIP-based image classification,~\citep{menon2022visual,yang2023language,pratt2023does} use GPT-3 to generate natural language attributes or captions for each class, and then classify images based on such information. Despite the success of these methods, they all suffer from large inference cost due to the use of LLM at test time. More critically, the LLM-generated texts are not necessarily beneficial, since they might be noisy and irrelevant to the considered task (see one example in Fig.~\ref{fig:schema}(a)).

To address the two issues, we propose a new prompt learning method that is boosted by task-relevant language priors but does not incur any LLM cost at test time. The high-level idea is to learn prompts via distillation of input-adapted textual knowledge, which is especially useful to recognize under-represented visual concepts. Specifically, for classification of \emph{object-centric images}, we follow~\citep{pratt2023does} to first query GPT-3 for a set of natural language prompts that describe each class. While for more complex tasks like VQA, we use human-generated image captions that can depict \emph{multi-object images} with object interactions in cluttered background. More importantly, for both cases, we \emph{learn} to aggregate the collected reference prompts into a single prompt embedding, which is optimized by the CLIP reward to have high similarity with input image.
% In other words, we learn an input-adapted prompt aggregator.
This allows us to obtain a condensed prompt embedding that is image-aligned, ruling out redundant and irrelevant information,~\eg,~ignoring the prompt elements about headlights for an image of rear facing car.
Finally, we jointly train a prompt generator conditioned on input image to generate a prompt embedding with two objectives: 1) staying close to the aggregated embedding (\ie,~distillation from the condensed textual knowledge), 2) minimizing the task loss (\ie,~downstream adaptation).

% Note our method inherits prompt learning's parameter efficiency, a merit in low-data regimes.
Fig.~\ref{fig:schema} illustrates our ``aggregate-and-adapt'' method for prompt learning. Note prompt aggregation and distillation is only required for the learning stage. At test time, we will discard the aggregator and use the learned prompt generator as a standalone module. This leads to compute-efficiency when compared to prior works~\citep{yang2021empirical,su2022language,menon2022visual,yang2023language,pratt2023does} since we entirely eliminate the LLM-induced inference cost.
Our generated \textbf{A}ggregate-and-\textbf{A}dapted \textbf{P}rompt \textbf{E}mbedding (or \textbf{\method}) prove highly effective on various downstream vision-language tasks. We show \method is a new state-of-the-art for few-shot image classification on 11 datasets, under different OOD generalization settings. \method can also generalize zero-shot to tasks like image-to-text retrieval, image captioning and VQA. When finetuned on these tasks, \method achieves even better performance than SOTA vision-language models (\eg,~MAGMA~\citep{eichenberg2022magma}) whose entire image and text networks are fine-tuned at large cost.

To summarize, our \textbf{main contributions} are:
\vspace{-0.1in}
\begin{itemize}[leftmargin=20pt]
\setlength{\itemsep}{0pt}
\item A new prompt learning method that distills the textual knowledge from human- or LLM-generated natural language prompts to improve the downstream generalization of CLIP.
\item Our learned \method achieves compelling performance on various downstream vision-language tasks, including image-to-text retrieval, few-shot classification, image captioning and VQA.
\item We offer insightful findings that \method is especially helpful when there are under-represented concepts in few-shot and OOD settings or ambiguous visual cues in non-canonical image views. \method learning is also data-efficient and scales better than baselines with LLM model size.
\end{itemize}
\vspace{-0.1in}

\section{Related Work}

\paragraph{Vision-language models.} Large-scale vision-language models achieve remarkable performance on a variety of downstream tasks. One learning paradigm is based on generative encoder-decoder models, which allows a sequence-to-sequence learning format that can connect visual data to free-form language prompts~\citep{alayrac2022flamingo,eichenberg2022magma,PaLI,mm1}. Recent works like LiMBeR~\citep{merullo2023linearly} show it is also possible for an LLM to operate on simple linear mappings of visual features. Another learning paradigm is based on contrastive learning with image-text pairs, which is popularized by CLIP~\citep{radford2021learning} and numerous follow-ups~\citep{jia2021scaling,li2022supervision,yao2022filip,rao2021denseclip,singh2022flava,yuan2021florence,10377550,lavoie2024modeling}. However, both categories of vision-language models (\eg,~CLIP and generative PaLI~\citep{PaLI}) are found to struggle with special visual concepts or domains. In this paper, we focus on CLIP and improve its downstream generalization, especially under few-shot and OOD settings, via distillation of language priors. Nevertheless, our approach can be applied to other vision-language models, which we leave as future work.

\paragraph{Prompt learning} is a parameter-efficient yet effective framework to finetune CLIP even in low-data settings. Most prompt learning methods learn text prompt vectors~\citep{zhou2021coop,zhou2022cocoop} in place of hand-written sentence prompts. Other methods show the possibility of learning prompts in the image space~\citep{jia2022visual}, or in both image and text spaces~\cite{zang2022unified,khattak2023maple}. To reduce overfitting to seen classes during prompt learning, recent works focus on new class feature synthesis~\citep{wang2023improving,zang2024overcoming}, improved optimization~\citep{shu2023clipood,lee2023rpo} and regularization~\citep{khattak2023self} strategies. More related to our approach are~\citep{yao2023visual,bulat2023lasp,zhuBeier2023} that align the learned prompts with hand-written prompts, with the goal of not forgetting the text knowledge from human input. These methods can be interpreted as a way of knowledge distillation from only short prompt templates. We will show the distillation from such prompt templates is suboptimal when compared to distillation from natural language prompts using LLMs.

\paragraph{Leveraging language in vision tasks.} There is a long line of works on leveraging language to aid vision or multimodal tasks. One family of methods rely on external natural language datasets to retrieve text knowledge of image categories. For example, \citep{bujwid-sullivan-2021,shen2022k} show improvements on ImageNet classification using the class descriptions retrieved from WordNet~\citep{Miller95} and ImageNet-Wiki~\citep{bujwid-sullivan-2021}. More recent works use LLMs to generate text for downstream tasks. GPT-3 is used in~\citep{yang2021empirical,su2022language} to help with the VQA and image captioning tasks. GPT-3 is also used in~\citep{menon2022visual,yang2023language,pratt2023does} to generate class-wise attributes or captions for CLIP-based classification. Unfortunately, all these prior works suffer from the noisy text that may be task-irrelevant. In this paper, we learn to adapt LLM-generated text to the target task, but without incurring any LLM-induced inference cost.

\section{Method}
Our ``aggregate-and-adapt'' method for prompt learning consists of three key components: 1) generating natural language prompts per image or class, 2) learning an aggregated prompt embedding that aligns with input image and 3) learning to generate Aggregate-and-Adapted Prompt Embedding (\method) for downstream tasks. In the following, we provide the details for each component. Note our method is based on CLIP~\citep{radford2021learning}, but it can be easily applied to other CLIP-like vision-language models.

\subsection{Generating Natural Language Prompts}
\label{sec:gen_nlp}

\paragraph{Prompt engineering.} For CLIP-based image classification, the standard approach requires a set of hand-written prompts,~\eg,~``a $\{\}$ in a video game'' and ``a dark photo of a $\{\}$'', which are completed with the class name. However, this is costly because one needs to hand-construct a different set of prompt templates for each dataset (\eg,~80 for ImageNet in~\citep{radford2021learning}), and that requires excessive prior knowledge about the target domain. Moreover, such prompt templates lack the descriptive details for discriminating fine-grained classes.

\paragraph{LLM-generated prompts.} For image classification, we make use of the rich knowledge in LLMs to generate natural language prompts for a given class. One benefit of using LLMs is the ability to generate an arbitrary number of prompts, without relying on any domain knowledge. In particular, we follow the CuPL method~\citep{pratt2023does} and query GPT-3~\citep{brown2020language} for a prompt set in a scalable way. Specifically, GPT-3 is queried with a few \emph{LLM-prompts} such as ``Describe what a(n) $\{\}$ looks like'' and ``How can you identify a(n) $\{\}$?''. Then for each LLM-prompt, GPT-3 generates 10 reference prompts using a high temperature of 0.99 for diversity. Fig.~\ref{fig:schema}(a) shows some example prompts for a particular car model ``Jeep Compass SUV 2012''. Note how the reference prompts specify the car's discriminating characteristics in its sleek exterior. For the 11 classification datasets considered in our work, we follow the full generation setting in ~\citep{pratt2023does}: for each dataset, we use a different set of LLM-prompts (between 2 to 9), resulting in 20-90 reference prompts generated for each class.

\begin{figure}[!t]
% \vskip -0.1in
\begin{center}
\centerline{\includegraphics[width=1.05\linewidth]{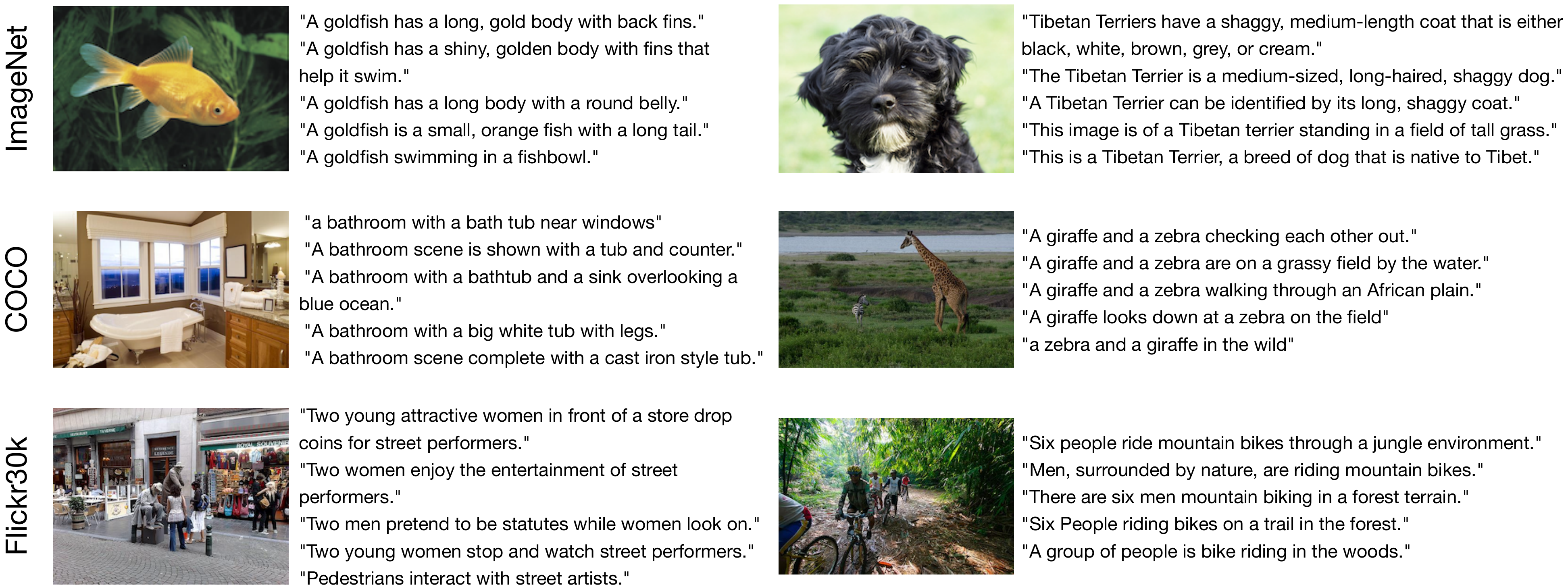}}
\caption{LLM-generated image prompts for ImageNet categories, and the hand-constructed image captions on COCO and Flickr30k datasets. Note ImageNet mainly contains \textbf{object-centric images} with relatively clean background, and the LLM-generated image prompts can describe distinct characteristics of the given classes. While COCO and Flickr30k contain \textbf{multi-object images} with cluttered background, and the hand-constructed captions can represent varying object relations.}
\label{fig:example_prompts}
\end{center}
% \vskip -0.1in
\end{figure}

\paragraph{Human-generated image captions.} As illustrated in Fig.~\ref{fig:example_prompts}, the LLM-generated class-wise prompts are mainly suited for the classification task, where there are often object-centric images with clean background. For more complex vision-language tasks like VQA, deeper understanding is required for multi-object scenes with varying object interactions and cluttered background. To capture the textual knowledge for describing multi-object images, we use their image captions available from image-text datasets as a source of natural language prompts. Here we use COCO dataset~\citep{LinMBHPRDZ14} that consists of 5 human-annotated captions per image. It will be shown that our prompt embedding learned on COCO suffices to generalize to three difficult vision-language tasks (image-to-text retrieval, image captioning and VQA) with varying data distributions.

\subsection{Input-Adapted Prompt Aggregator}
The LLM-generated prompts have one notable issue: they are not necessarily a good representation of input image. For example in Fig.~\ref{fig:schema}(a), it is inaccurate to describe the input image of a silver Jeep SUV as a red one in the last prompt, whereas other prompts are more relevant. Obviously, it would be detrimental to directly use the noisy prompts to supervise the following learning stage. Another issue with both the LLM- and human-generated prompts is that they are highly redundant with repeated information. To find a better supervisory signal, we propose to first aggregate the reference prompts into an image-aligned, condensed ``summary''. Such prompt summary is expected to have filtered noise as well as reduced redundancy. 

Given $n$ generated prompts, we first use the text encoder of CLIP to obtain the prompt embeddings $\mP = [\vp_1, \vp_2, \ldots, \vp_n]$. Then we learn an adaptive prompt aggregator to condense $\mP$ into $m$ ($\ll n$) embeddings of the same size. Ideally, the aggregation should be invariant to permutations of $\mP$, and scales as $\mathcal{O}(n+m)$. Here we set $m=1$ for efficiency concerns. The aggregated result, a single prompt embedding, is denoted as $\vp^a$. One simple aggregator that has the properties of permutation invariance and high efficiency is based on just averaging $\mP$ into $\bar{\vp}$. Simple averaging is widely used in prior works~\citep{menon2022visual,pratt2023does}. However, the mean $\bar{\vp}$ would still be compromised by the irrelevant information in $\mP$. Here we introduce an attention-based aggregator which allow us to align the aggregated $\vp^a$ with input image. At the same time, our prompt aggregator remains efficient and permutation invariant.

\begin{figure}[!t]
\vskip -0.15in
\begin{center}
\centerline{\includegraphics[width=1.0\linewidth]{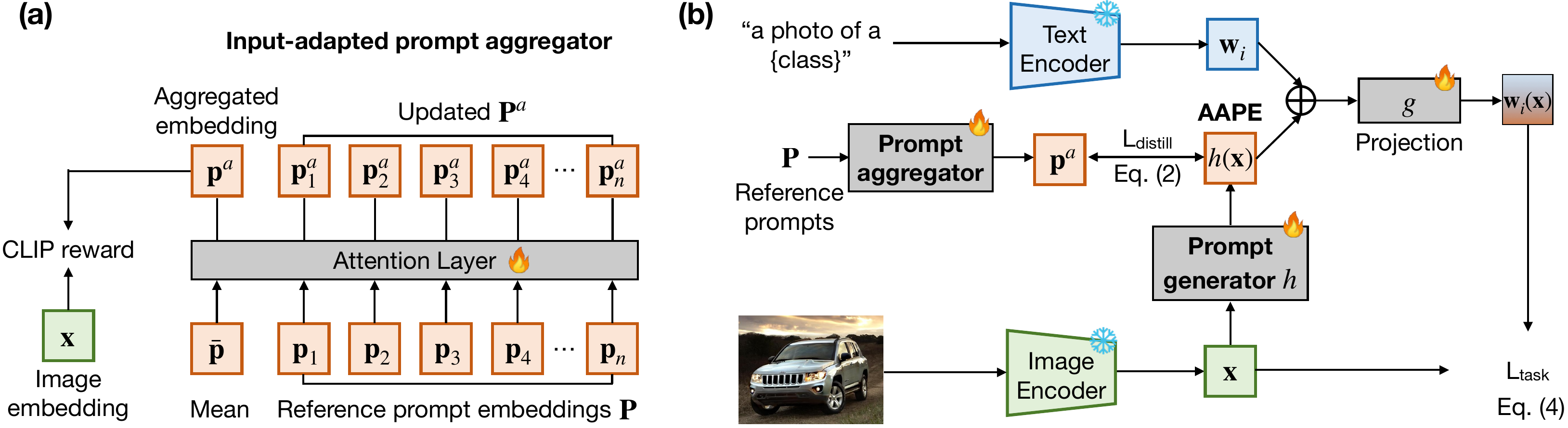}}
% \vskip -0.1in
\caption{\textbf{(a)} Input-adapted prompt aggregator which aggregates the embeddings of reference prompts $\mP$ into an image-aligned, condensed prompt embedding $\vp^a$ based on CLIP reward. \textbf{(b)} Instantiation of our prompt learning approach for image classification. The CLIP model is kept frozen.}
\label{fig:aggregator_classfication_scheme}
\end{center}
\vskip -0.2in
\end{figure}

Fig.~\ref{fig:aggregator_classfication_scheme}(a) shows the architecture of our input-adapted prompt aggregator based on just one attention layer. The attention layer takes as input the reference prompts $\mP$ and a learnable prompt embedding (initialized as $\bar{\vp}$). Then all the embeddings are updated as follows:
\begin{equation}
    \left[\vp^a,\mP^a\right] = \texttt{AttentionLayer}\left(\left[\bar{\vp},\mP\right]\right),
\label{eq1}
\end{equation}
where $\vp^a$ is the desired prompt aggregation. The attention layer consists of standard multi-head cross-attention and feed-forward networks together with LayerNorm~\citep{ba2016layer}.

To make $\vp^a$ semantically related to the input image (with embedding $\vx$), we optimize our prompt aggregator using the CLIP reward~\citep{clipscore} in form of $\texttt{CLIP-S}(\vx,\vp^a)=s\cdot \max(\cos(\vx,\vp^a),0)$. The CLIP reward allows $\vp^a$ to selectively blend image-related prompts through the attention mechanism. Fig.~\ref{fig:prompt_score} in Appendix~\ref{sec:appendix_ablation_aggregator} confirms that redundant or irrelevant reference prompts tend to have low attention scores, hence they are suppressed during prompt aggregation.

\subsection{Learning \method}
\label{sec:AAPE}

The per-image prompt aggregation $\vp^a$ offers useful textual knowledge to supervise the following prompt learning stage. In this section, we elaborate how to improve prompt learning by distilling the aggregated text knowledge from $\vp^a$. Note CLIP is kept frozen during prompt learning.

As a key innovation of this paper, we propose to train a prompt generator $h$ that directly generates the prompt embedding $h(\vx)$ conditioned on image features $\vx$. We parameterize $h$ as a lightweight network with two fully connected layers and ReLU nonlinearity. $h$ is trained using a distillation loss $\mathcal{L}_{\text{distill}}$ for knowledge distillation from $\vp^a$, as well as a task loss $\mathcal{L}_{\text{task}}$ for downstream adaptation. We call such learned $h(\vx)$ as Aggregate-and-Adapted Prompt Embedding (\method). \method can also be viewed as an \textit{image captioning embedding} in the latent space, since useful text knowledge is distilled in \method. In the following, we detail the two training losses.

\paragraph{Distillation loss} is simply defined as the Euclidean distance between $h(\vx)$ and $\vp^a$:
\begin{equation}
    \mathcal{L}_{\text{distill}} = \|h(\vx)-\vp^a\|_{2}^{2}.
\label{eq2}
\end{equation}

\paragraph{Task loss -- image classification.} Besides distilling the textual knowledge from $\vp^a$, $h(\vx)$ should adapt to the downstream task too. Here we start with the instantiation of adapting $h(\vx)$ to the most studied task of image classification.

% which are then matched to the image features $\vx$ to perform classification
Note existing prompt learners for classification (\eg,~\citep{zhou2021coop,zhou2022cocoop}) learn \textit{individual word tokens} $\{\vv_l\}_{l=1}^L$ in a prompt, and then combine the learned tokens with class name embeddings $\vc_{i\in [1,\ldots,C]}$ to obtain the full prompt. By contrast, we directly generate a full prompt embedding $h(\vx)$ without token-wise prediction. For classification, we simply combine $h(\vx)$ and the embedding of a prompt template ``a photo of a \{\texttt{class}\}'' to act as the classifier weights (to be matched to image features $\vx$).

Fig.~\ref{fig:aggregator_classfication_scheme}(b) shows the overall prompt learning framework. For the prompt template filled with the $i$-th class name, we use the text encoder of CLIP to obtain the template embedding $\vw_i \in \mathbb{R}^{d}$. Next, we concatenate $\vw_{i}$ and our $h(\vx) \in \mathbb{R}^{d}$ followed by a projection $g$, giving $\vw_{i}(\vx) = g\left([\vw_{i}^{\top},h(\vx)^{\top} ]^{\top}\right)$. Note $\vw_{i}(\vx)$ does not involve any prompt engineering effort. We rely on $\vw_{i}$ to mainly encode the class name, while $h(\vx)$ enriches that with input-adapted class descriptions in the latent space.

We parameterize $g\!\!:\mathbb{R}^{2d} \rightarrow \mathbb{R}^{d}$ by one fully connected layer with ReLU nonlinearity. The nonlinearity is important since it ensures $\vw_{i}(\vx)$ is not trivially equivalent to the linear combination of $\vw_{i}$ and $h(\vx)$, which will ignore $h(\vx)$ if we match the linear combination to $\vx$ for classification. The classification probability is given as:
\begin{equation}
    p\left(y=c \mid \vx\right)=\frac{\exp \left(\cos \left(\vx,\vw_{c}(\vx)\right) / \tau\right)}{\sum_{i=1}^C \exp \left(\cos \left(\vx,\vw_{i}(\vx)\right) / \tau\right)},
\label{eq3}
\end{equation}
where $C$ is the total number of classes, $\tau$ and $\cos(\cdot,\cdot)$ denote the temperature and cosine similarity. Appendix~\ref{sec:appendix_other_classification_framework} (Table~\ref{tb:hx_only_classification}) compares with an $h(\vx)$-only baseline for classification, without combining $\vw_{i}$ or using projection $g$. Results show our default classification framework achieves solid gains over the $h(\vx)$-only baseline.

Finally, we arrive at the overall loss function to train the prompt generator $h$ together with projection $g$ for image classification:
\begin{equation}
    \mathcal{L}=\lambda \mathcal{L}_{\text{distill}} + \mathcal{L}_{\text{task}}, \;\;\; \text{where} \;\; \mathcal{L}_{\text{task}}= - \log p\left(y=c \mid \vx\right),
\label{eq4}
\end{equation}
and $\lambda=5$ is a weighting parameter. Table~\ref{tb:lambda_ablation} in Appendix~\ref{sec:appendix_hyperparameter_cost} provides the sensitivity analysis for $\lambda$, which shows performance is quite robust to the $\lambda$ value in a wide range.

Note during testing we only use the prompt generator $h$, without querying LLM or using the prompt aggregator anymore. This removes the LLM-induced inference cost as required in~\citep{yang2021empirical,su2022language,menon2022visual,yang2023language,pratt2023does}, shifting such cost into our learning stage.

\paragraph{Task loss -- beyond classification.} Will the textual knowledge in $h(\vx)$ or \method benefit other vision-language tasks? We consider three tasks beyond classification: \textbf{image-to-text retrieval}, \textbf{image captioning} \& \textbf{VQA}. Note these tasks involve multi-object images as mentioned in Section~\ref{sec:gen_nlp}. Hence we use COCO image captions that contain textual descriptions of object relations. For all the three tasks, we train both the prompt aggregator and prompt generator $h$ on 5 COCO captions per image. The same $\mathcal{L}_{\text{distill}}$ in Eq.~(\ref{eq2}) is used, while $\mathcal{L}_{\text{task}}$ is the CLIP loss.

For image-text retrieval, we simply use \method as the query and evaluate its zero-shot text retrieval performance on Flickr30k dataset~\citep{young-etal-2014-image}. We also evaluate the finetuning performance on Flickr30k when the prompt generator $h$ is finetuned using $\mathcal{L}_{\text{distill}}$ and the corresponding retrieval loss $\mathcal{L}_{\text{task}}$.

For image captioning and VQA tasks, we conjecture that the textual knowledge encoded in \method will be especially useful when the visual cues are confusing or missing. To test this hypothesis, we use the COCO-trained \method in a straightforward manner. Concretely, we follow the LiMBeR baseline in~\citep{merullo2023linearly} which linearly transforms the CLIP image representation into a sequence of prompt embeddings that an LLM can process. Then \method is appended to the prompt sequence as a ``prefix'' to offer rich language priors. We similarly evaluate performance of the zero-shot or finetuned prompt generator $h$ on downstream datasets. For finetuning, $h$ is optimized using $\mathcal{L}_{\text{distill}}$ and the corresponding task loss $\mathcal{L}_{\text{task}}$ (captioning or VQA).

\section{Experimental Setup and Datasets}
\label{sec:setup_datasets}

\subsection{Few-shot Image Classification}

\paragraph{Datasets.} We use 11 datasets: ImageNet~\citep{deng2009imagenet}, Caltech101~\citep{fei2004learning}, OxfordPets~\citep{parkhi2012cats}, StanfordCars~\citep{krause20133d}, Flowers102~\citep{nilsback2008automated}, Food101~\citep{bossard2014food}, FGVC-Aircraft~\citep{maji2013fine}, SUN397~\citep{xiao2010sun}, UCF101~\citep{soomro2012ucf101}, DTD~\citep{cimpoi2014describing} and EuroSAT~\citep{helber2019eurosat}. These datasets cover a wide range of generic objects and scenes, fine-grained object classes, as well as special domains with textural and satellite images. The various visual concepts in these datasets are perfect to test whether and when the textual knowledge in LLM will help. We further evaluate domain generalization on ImageNetV2~\citep{recht2019imagenet}, ImageNet-Sketch~\citep{wang2019learning}, ImageNet-A~\citep{hendrycks2021natural} and ImageNet-R~\citep{hendrycks2021many}, which have different types of domain shift from ImageNet.

\paragraph{Implementation.} We follow the prompt learning details in~\citep{zhou2021coop}, including the CLIP vision backbone (ViT-B/16), learning rate schedule and the number of epochs for each dataset. Appendix~\ref{sec:appendix_hyperparameter_cost} (Table~\ref{tb:inference_cost}) provides a detailed analysis of the compute cost measured on Nvidia V100 GPU, where all prompt learners are evaluated for fair efficiency comparisons.

\paragraph{Evaluations.} We follow the few-shot evaluation protocol in~\citep{radford2021learning}, using 1, 2, 4, 8 and 16 shots per class for training (default 16), and the full testset for evaluation. Two OOD generalization settings are considered as in~\citep{zhou2021coop}. 1) Generalization from base to new classes within one dataset,~\ie,~training on the base class split but testing on both base and new class splits. This helps evaluate the ID and OOD performance under a class-incremental domain shift. We follow~\citep{xian2017zero} to also measure the harmonic mean of base and new class accuracies to quantify the ID and OOD performance trade-off. 2) Domain generalization where one trains on ImageNet (with 16 shots) and evaluates on four ImageNet variants. For all experiments, we report results as an average over three random seeds.

\subsection{Vision-Language Understanding and Generation Tasks}

As mentioned in Section~\ref{sec:AAPE}, we perform prompt learning on COCO dataset~\citep{LinMBHPRDZ14} before evaluation on three vision-language tasks. For the task of image-to-text retrieval, we use the same CLIP vision backbone VIT-L/14 as in~\citep{radford2021learning,lavoie2024modeling} for fair comparisons. We show both zero-shot and finetuned results (Recall@K) on Flickr30k~\citep{young-etal-2014-image}.

For captioning and VQA tasks, we follow LiMBeR~\citep{merullo2023linearly} to use the same language model and CLIP vision backbone (RN50x16). We evaluate on image captioning datasets COCO and NoCaps~\citep{agrawal2019nocaps}. Zero-shot and finetuned results are reported in terms of CIDEr-D~\citep{VedantamZP15}, CLIPScore, and Ref-CLIPScore~\citep{clipscore}. For VQA, we prompt the model with the ``[image] Q: [q] A:'' format. The generation is truncated to the length of the longest ground truth answer. For evaluation, we use the VQA2 dataset~\citep{balanced_vqa_v2} and follow the few-shot setting in~\citep{eichenberg2022magma} to report accuracy metric for every K-shots.

Appendix~\ref{sec:appendix_hyperparameter_cost} (Table~\ref{tb:inference_cost}) shows the high efficiency with the straightforward use of \method for tasks beyond classification. When compared to the SOTA fully fine-tuned model MAGMA~\citep{eichenberg2022magma}, \method is about 2.8/1.2 times faster for training/inference on Nvidia A100 GPU.

\section{Results}

\subsection{Image-to-Text Retrieval}

We use this task to verify if \method can act as a meaningful image captioning embedding, which is learned to distill image-aligned text knowledge from available image captions. We do not aim to push for state-of-the-art performance for the retrieval task.
Table~\ref{tb:text_retrieval} shows that we can indeed achieve strong training and finetuning performance on COCO and Flickr30k datasets, respectively. This indicates our prompt learning method is competent with producing high-quality text or captioning embedding that can successfully fulfill the task at hand.
It is also worth noting that our \method learned on COCO can perform zero-shot retrieval on Flickr30k, obtaining competitive results with SOTA zero-shot models (\eg,~CLIP and SigLIP). This further demonstrates the good generalization capability of \method over different data distributions.

\begin{table*}[!t]
\caption{\textbf{Image-to-Text Retrieval:} zero-shot and finetuned results (Recall@K) on Flickr30k dataset. Note our \method is learned on the COCO dataset.}
\label{tb:text_retrieval}
% \vskip -0.1in
\begin{center}
\resizebox{0.9\linewidth}{!}{
\begin{tabular}{clcccccc}
\toprule
& & \multicolumn{3}{c}{COCO} & \multicolumn{3}{c}{Flickr30k} \\
\cmidrule(lr){3-5}
\cmidrule(lr){6-8}
& & R@1 & R@5 & R@10 & R@1 & R@5 & R@10 \\
\midrule
\multirow{4}{*}{\rotatebox[origin=c]{90}{\shortstack[c]{Train or\\Finetune}}} & Unicoder-VL~\citep{Li2019UnicoderVLAU} & 62.3 & 87.1 & 92.8 & 86.2 & 96.3 & 99.0\\
& Oscar~\citep{Xiujun2020} & 73.5 & 92.2 & 96.0 & - & - & -\\
& ERNIE-ViL~\citep{Yu2020ERNIEViLKE} & - & - & - & 88.7 & 98.0 & 99.2\\
& \method & \textbf{76.7}\textsubscript{$\pm$0.1} & \textbf{94.5}\textsubscript{$\pm$0.1} & \textbf{97.4}\textsubscript{$\pm$0.1} & \textbf{94.2}\textsubscript{$\pm$0.2} & \textbf{99.3}\textsubscript{$\pm$0.1} & \textbf{99.7}\textsubscript{$\pm$0.1}\\ \midrule
\multirow{4}{*}{\rotatebox[origin=c]{90}{Zero-shot}} & CLIP~\citep{radford2021learning} & 58.4 & 81.5 & 88.1 & 88.0 & 98.7 & 99.4\\
& SigLIP~\citep{10377550} & 65.4 & 85.1 & 91.1 & 91.5 & 98.1 & 99.4\\
& Llip~\citep{lavoie2024modeling} & \textbf{68.1} & \textbf{87.6} & \textbf{92.5} & 93.2 & 99.0 & 99.4\\
& BLIP-2~\citep{li2023blip2} & - & - & - & \textbf{96.9} & \textbf{100.0} & \textbf{100.0}\\
& \method & - & - & - & 90.8\textsubscript{$\pm$0.2} & 98.6\textsubscript{$\pm$0.1} & 99.4\textsubscript{$\pm$0.1}\\ 
\bottomrule
\end{tabular}
}
\end{center}
% \vskip -0.2in
\end{table*}

\subsection{Few-shot Image Classification}

\paragraph{Base-to-new class generalization.} Table~\ref{tb:base2new} compares \method with two categories of prompt learning methods: 1) CoOp~\citep{zhou2021coop}, CoCoOp~\citep{zhou2022cocoop}, MaPLe~\citep{khattak2023maple}, CLIPood~\citep{shu2023clipood}, PromptSRC~\citep{khattak2023self} and OGEN~\citep{zang2024overcoming}. These methods learn prompt vectors without using any text-based knowledge, but heavily rely on advanced optimization and regularization strategies to improve generalization. Our \method, on the other hand, outperforms by distilling the textual knowledge from LLMs. Notably, on average (across 11 datasets), \method achieves better classification accuracies than the previous SOTA approach OGEN for both base and new classes, setting a new SOTA mean accuracy 80.97\% (vs. 80.34\%). 2) ProGrad~\citep{zhuBeier2023}, KgCoOp~\citep{yao2023visual} and LASP-V~\citep{bulat2023lasp}. These methods choose to align the learned prompts with hand-written ones like ``a dark photo of a $\{\texttt{class}\}$'', hence distilling knowledge from only basic, non-descriptive templates. This proves less effective than our LLM-derived natural language priors.

\begin{table*}[t!]
\begin{center}
\vskip -0.2in
\caption{\textbf{Few-shot classification in the base-to-new class generalization setting}. OGEN denotes the OGEN+PromptSRC variant. Our \method follows CuPL to query an LLM to obtain natural language prompts, but further learns from those prompts. H: Harmonic mean of base and new class accuracies.}
\label{tb:base2new}
% \tablestyle{+1pt}{0.9}
\addtolength{\tabcolsep}{-3pt}
\resizebox{\linewidth}{!}{
\begin{tabular}{lc | cccccc | ccc | cc }
\toprule
 & \multicolumn{1}{c@{}}{} & \multicolumn{6}{c}{\textbf{Prompt learning without language priors}} & \multicolumn{3}{c}{\textbf{Human-generated prompts}} & \multicolumn{2}{c}{\textbf{LLM-based}} \\
 \cmidrule(lr){3-8} \cmidrule(lr){9-11} \cmidrule(lr){12-13}
 & \multicolumn{1}{c@{}}{} & CoOp & CoCoOp & MaPLe & CLIPood & PromptSRC & \multicolumn{1}{c@{}}{OGEN} &ProGrad & KgCoOp & \multicolumn{1}{c@{}}{LASP-V} & CuPL & \method \\
\midrule
\multirow{3}{*}{\shortstack[l]{Avg across\\ 11 datasets}} & Base & 82.69 & 80.47 & 82.28 & 83.90 & 84.26 & 84.17 & 82.48 & 80.73 & 83.18 & 74.31 & \textbf{84.72}\textsubscript{$\pm$0.18}\\
& New & 63.22 & 71.69 & 75.14 & 74.50 & 76.10 & 76.86 & 70.75 & 73.60 & 76.11 & 75.25 & \textbf{77.54}\textsubscript{$\pm$0.29}\\
& H & 71.66 & 75.83 & 78.55 & 78.90 & 79.97 & 80.34 & 76.16 & 77.00 & 79.48 & 74.78 & \textbf{80.97}\textsubscript{$\pm$0.19}\\
\midrule
\multirow{3}{*}{ImageNet} & Base & 76.47 & 75.98 & 76.66 & 77.50 & 77.60 & 77.50 & 77.02 & 75.83 & 76.25 & 75.05 & \textbf{78.10}\textsubscript{$\pm$0.11}\\
& New & 67.88 & 70.43 & 70.54 & 70.30 & 70.73 & 70.97 & 66.66 & 69.96 & 71.17 & 68.43 & \textbf{71.98}\textsubscript{$\pm$0.14}\\
& H & 71.92 & 73.10 & 73.47 & 73.70 & 74.01 & 74.09 & 71.46 & 72.78 & 73.62 & 71.59 & \textbf{74.92}\textsubscript{$\pm$0.12}\\
\midrule
\multirow{3}{*}{Caltech101} & Base & 98.00 & 97.96 & 97.74 & \textbf{98.70} & 98.10 & 98.32 & 98.02 & 97.72 & 98.17 & 98.24 & 98.34\textsubscript{$\pm$0.07}\\
& New & 89.81 & 93.81 & 94.36 & 94.60 & 94.03 & 94.76 & 93.89 & 94.39 & 94.33 & 94.34 & \textbf{94.79}\textsubscript{$\pm$0.09}\\
& H & 93.73 & 95.84 & 96.02 & \textbf{96.60} & 96.02 & 96.50 & 95.91 & 96.03 & 96.21 & 96.25 & 96.53\textsubscript{$\pm$0.07}\\
\midrule
\multirow{3}{*}{OxfordPets} & Base & 93.67 & 95.20 & 95.43 & 95.70 & 95.33 & 95.96 & 95.07 & 94.65 & 95.73 & 95.30  & \textbf{96.89}\textsubscript{$\pm$0.12}\\
& New & 95.29 & 97.69 & 97.76 & 96.40 & 97.30 & 97.48 & 97.63 & 97.76 & 97.87 & 97.74 & \textbf{98.02}\textsubscript{$\pm$0.16}\\
& H & 94.47 & 96.43 & 96.58 & 96.00 & 96.30 & 96.71 & 96.33 & 96.18 & 96.79 & 96.50 & \textbf{97.45}\textsubscript{$\pm$0.13}\\
\midrule
\multirow{3}{*}{\shortstack[l]{Stanford \\ Cars}} & Base & 78.12 & 70.49 & 72.94 & \textbf{78.60} & 78.27 & 77.59 & 77.68 & 71.76 & 75.23 & 68.88 & 77.51\textsubscript{$\pm$0.39}\\
& New & 60.40 & 73.59 & 74.00 & 73.50 & 74.97 & 75.17 & 68.63 & 75.04 & 71.77 & 75.09 & \textbf{77.37}\textsubscript{$\pm$0.56}\\
& H & 68.13 & 72.01 & 73.47 & 75.90 & 76.58 & 76.38 & 72.88 & 73.36 & 73.46 & 71.85 & \textbf{77.44}\textsubscript{$\pm$0.42}\\
\midrule
\multirow{3}{*}{Flowers102} & Base & 97.60 & 94.87 & 95.92 & 93.50 & \textbf{98.07} & 97.34 & 95.54 & 95.00 & 97.17 & 77.79 & 97.81\textsubscript{$\pm$0.19}\\
& New & 59.67 & 71.75 & 72.46 & 74.50 & 76.50 & 77.67 & 71.87 & 74.73 & 73.53 & 78.10 & \textbf{78.75}\textsubscript{$\pm$0.31}\\
& H & 74.06 & 81.71 & 82.56 & 82.90 & 85.95 & 86.39 & 82.03 & 83.65 & 83.71 & 77.94 & \textbf{87.25}\textsubscript{$\pm$0.21}\\
\midrule
\multirow{3}{*}{Food101} & Base & 88.33 & 90.70 & 90.71 & 90.70 & 90.67 & 90.69 & 90.37 & 90.50 & 91.20 & 90.56 & \textbf{91.82}\textsubscript{$\pm$0.08}\\
& New & 82.26 & 91.29 & 92.05 & 91.70 & 91.53 & 91.68 & 89.59 & 91.70 & 91.90 & 91.86 & \textbf{92.66}\textsubscript{$\pm$0.11}\\
& H & 85.19 & 90.99 & 91.38 & 91.20 & 91.10 & 91.19 & 89.98 & 91.09 & 91.54 & 91.21 & \textbf{92.24}\textsubscript{$\pm$0.09}\\
\midrule
\multirow{3}{*}{\shortstack[l]{FGVC \\ Aircraft}} & Base & 40.44 & 33.41 & 37.44 & \textbf{43.30} & 42.73 & 41.26 & 40.54 & 36.21 & 38.05 & 33.29  & 41.46\textsubscript{$\pm$0.12}\\
& New & 22.30 & 23.71 & 35.61 & 37.20 & 37.87 & 40.26 & 27.57 & 33.55 & 33.20 & 37.60 & \textbf{40.37}\textsubscript{$\pm$0.28}\\
& H & 28.75 & 27.74 & 36.50 & 40.00 & 40.15 & 40.75 & 32.82 & 34.83 & 35.46 & 35.31 & \textbf{40.91}\textsubscript{$\pm$0.14}\\
\midrule
\multirow{3}{*}{SUN397} & Base & 80.60 & 79.74 & 80.82 & 81.00 & 82.67 & 82.57 & 81.26 & 80.29 & 80.70 & 73.39 & \textbf{82.93}\textsubscript{$\pm$0.14}\\
& New & 65.89 & 76.86 & 78.70 & 79.30 & 78.47 & 78.83 & 74.17 & 76.53 & 79.30 & 75.69 & \textbf{79.87}\textsubscript{$\pm$0.27}\\
& H & 72.51 & 78.27 & 79.75 & 80.20 & 80.52 & 80.65 & 77.55 & 78.36 & 80.00 & 74.52 & \textbf{81.37}\textsubscript{$\pm$0.18}\\
\midrule
\multirow{3}{*}{DTD} & Base & 79.44 & 77.01 & 80.36 & 80.80 & 83.37 & 83.75 & 77.35 & 77.55 & 81.10 & 62.45 & \textbf{83.97}\textsubscript{$\pm$0.45}\\
& New & 41.18 & 56.00 & 59.18 & 58.60 & 62.97 & 62.54 & 52.35 & 54.99 & 62.57 & 60.31 & \textbf{63.64}\textsubscript{$\pm$0.58}\\
& H & 54.24 & 64.85 & 68.16 & 67.90 & 71.75 & 71.60 & 62.45 & 64.35 & 70.64 & 61.36 & \textbf{72.41}\textsubscript{$\pm$0.39}\\
\midrule
\multirow{3}{*}{EuroSAT} & Base & 92.19 & 87.49 & 94.07 & \textbf{97.50} & 92.90 & 93.40 & 90.11 & 85.64 & 95.00 & 65.38 & 95.40\textsubscript{$\pm$0.33}\\
& New & 54.74 & 60.04 & 73.23 & 64.10 & 73.90 & 76.74 & 60.89 & 64.34 & \textbf{83.37} & 69.97 & 76.30\textsubscript{$\pm$0.86}\\
& H & 68.69 & 71.21 & 82.35 & 77.30 & 82.32 & 84.25 & 72.67 & 73.48 & \textbf{88.86} & 67.60 & 84.79\textsubscript{$\pm$0.63}\\
\midrule
\multirow{3}{*}{UCF101} & Base & 84.69 & 82.33 & 83.00 & 85.70 & 87.10 & 87.44 & 84.33 & 82.89 & 85.53 & 77.13 & \textbf{87.69}\textsubscript{$\pm$0.23}\\
& New & 56.05 & 73.45 & 78.66 & \textbf{79.30} & 78.80 & 79.28 & 74.94 & 76.67 & 78.20 & 78.58 & 79.21\textsubscript{$\pm$0.39}\\
& H & 67.46 & 77.64 & 80.77 & 82.40 & 82.74 & 83.16 & 79.35 & 79.65 & 81.70 & 77.85 & \textbf{83.23}\textsubscript{$\pm$0.31}\\
\bottomrule
\end{tabular}
}
\end{center}
% \vskip -0.2in
\end{table*}

Recall our prompt learning method is built on top of the CuPL approach~\citep{pratt2023does} to obtain the knowledge of GPT-3. Here we show both the LLM knowledge and our learning algorithm (that adaptively distills the knowledge) are indispensable. We first compare with the CuPL baseline that leverages LLM knowledge, but without any learning. Specifically, CuPL averages the LLM-generated prompts to perform zero-shot classification. This is contrasted with our method that \emph{learns} to aggregate the noisy prompts into \method which is then adapted for classification. Table~\ref{tb:base2new} shows such ``aggregate-and-adapt'' learning method leads to significant gains over the learning-free CuPL, especially for base classes.

\begin{figure}[!t]
\vskip -0.43in
\begin{center}
\centerline{\includegraphics[width=1.0\linewidth]{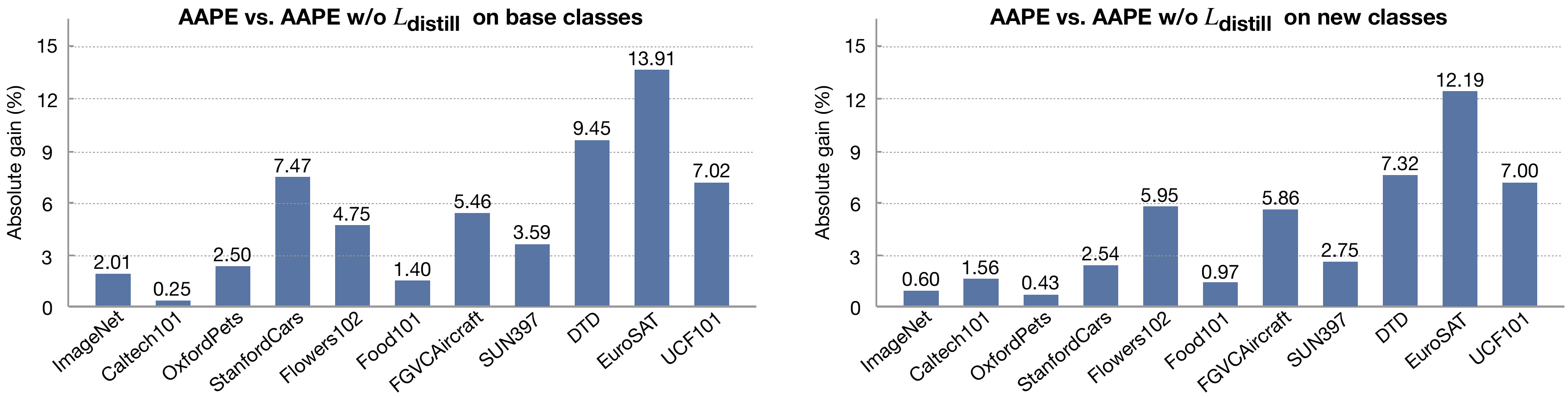}}
% \vskip -0.1in
\caption{\textbf{Quantifying the role of LLM knowledge (distilled with $\mathcal{L}_{\text{distill}}$) in prompt learning.} $\mathcal{L}_{\text{distill}}$ consistently improves the base and new class accuracies on 11 classification datasets.}
\label{fig:base2new_gain}
\end{center}
% \vskip -0.27in
\end{figure}

Next we compare with a variant of our approach, with only task loss $\mathcal{L}_{\text{task}}$ but no $\mathcal{L}_{\text{distill}}$ to distill LLM knowledge. This variant allows decoupling the contribution of LLM knowledge for prompt learning under a fair setting. Fig.~\ref{fig:base2new_gain} shows that using $\mathcal{L}_{\text{distill}}$ leads to consistent gains for both seen and unseen classes from all the considered datasets. The distilled textual knowledge makes an especially large impact for those fine-grained actions (UCF101) and visual classes (StanfordCars, Flowers102 and FGVCAircraft), which can be under-represented during both CLIP pretraining and prompt learning. Large gains are also observed for the special domains of textures (DTD) and satellite images (EuroSAT) with large distribution shift.

% This variant is similar to CoCoOp in that both learn prompts in an image-conditional way. But we learn the full prompt embedding instead of the embeddings of individual word tokens as in CoCoOp, hence improving the parameter efficiency. We achieve better average H metric: 76.23 vs. 75.83.

\begin{figure}[!t]
\vskip -0.2in
\begin{center}
\centerline{\includegraphics[width=1.0\linewidth]{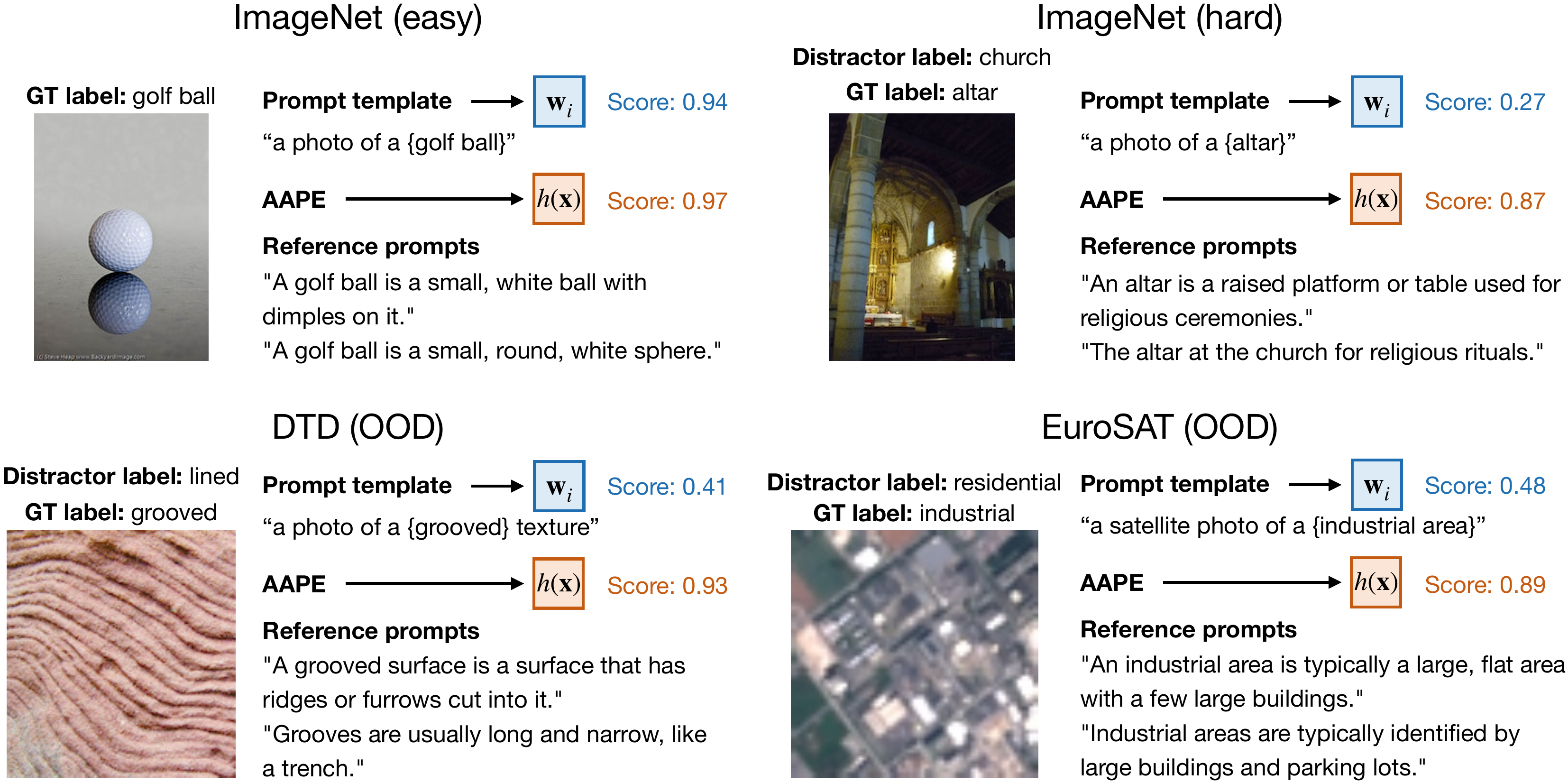}}
% \vskip -0.1in
\caption{\textbf{\method helps disambiguate the classification task}. To highlight the textual knowledge encoded in \method, we show some reference prompts generated by GPT-3. For both the prompt template and \method (before concatenation and projection), we measure their Cosine similarity score with the image. Note the similarity score can be small when using a basic prompt template to match the ``altar'' class instance on ImageNet. Indeed, in this non-canonical image view, the altar is small and the whole scene can be classified as the easily confused class of ``church''. Whereas \method is able to eliminate confusion by providing additional cues like altar ``is a raised table'' often at the location of ``church''. This results in increased image-text similarity. Similarly, the textual cues from \method are helpful for the OOD examples in special domains of DTD and EuroSAT.}
\label{fig:wi_AAPE}
\end{center}
\vskip -0.35in
\end{figure}

We further show how our distilled \method can disambiguate the classification task. Fig.~\ref{fig:wi_AAPE} shows our \method augments the basic prompt template with descriptive details for each image, as exemplified by the reference prompts. Such input-specific details are often helpful for non-canonical views (\eg,~hard cases on ImageNet) and OOD examples (\eg,~on DTD and EuroSAT), where the visual cues are either ambiguous or barely visible (hence low similarity between the image and basic template). Eventually, we use a projection network to blend textual information from the template and \method, resulting in increased image-text similarity.

\paragraph{More comparisons.} Table~\ref{tb:domain_generalization} in Appendix~\ref{sec:domain_generalization} includes \textbf{domain generalization} results. We see that \method is robust to different types of domain shift, outperforming prior works on 4 ImageNet variants.
Table~\ref{tb:ProText_ArGue} and Appendix~\ref{sec:ProText_ArGue} further show the advantage of \method over two recent prompt learning methods ProText~\citep{Khattak2024ProText} and ArGue-N~\citep{10657279}.

% \paragraph{More ablations.} Our prompt aggregator plays a key role of providing high-quality textual knowledge to supervise \method learning. Table~\ref{tb:aggregator_ablation} in Appendix~\ref{sec:aggregator_ablation} validates that our attention-based aggregator is better than alternatives using a different architecture or simple averaging.

\begin{figure}[!t]
\vskip -0.5in
\begin{center}
\centerline{\includegraphics[width=1.0\linewidth]{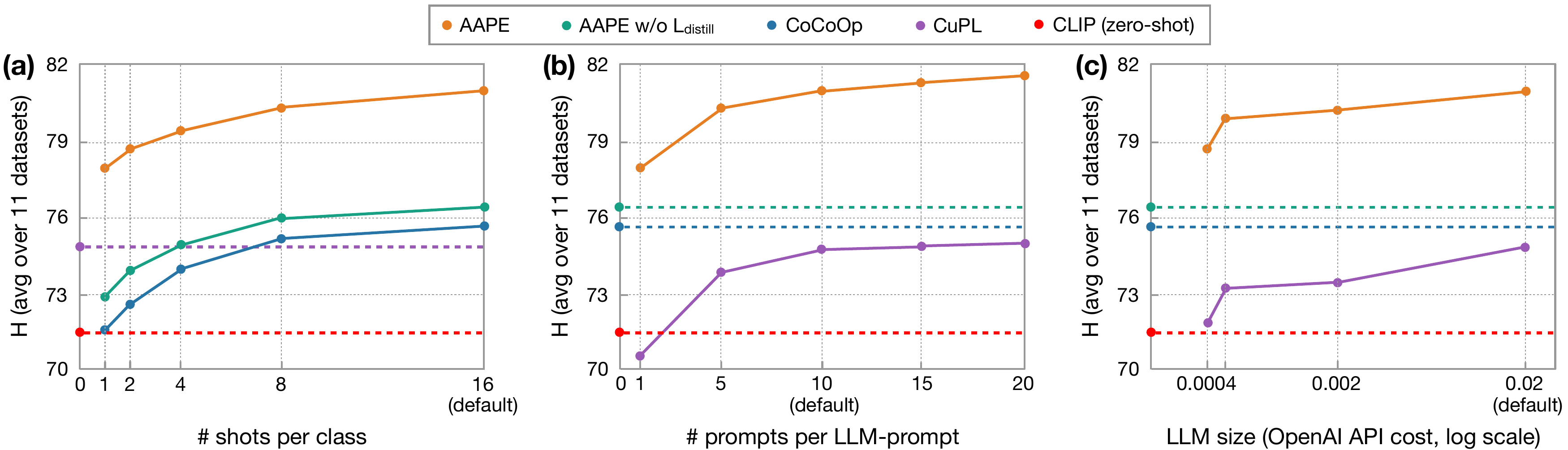}}
% \vskip -0.1in
\caption{\textbf{\method scales better with data (a-b) and LLM size (c) than alternatives}. Experiments are conducted under the base-to-new  generalization setting for few-shot classification. We measure the Harmonic mean (H) of base and new class accuracies. To adjust the total number of reference prompts per class to supervise \method learning, we vary the number of prompts generated by each LLM-prompt template. Four models of GPT-3 are considered: Ada, Babbage, Curie and Davinci.}
\label{fig:scaling_property}
\end{center}
\vskip -0.3in
\end{figure}

\textbf{Ablation studies.} Fig.~\ref{fig:scaling_property}(a) shows that \method learning is data-efficient. We see \method consistently outperforms two prompt learners that do not benefit from LLM's text knowledge,~\ie,~\method w/o $\mathcal{L}_{\text{distill}}$ and a similar baseline CoCoOp~\citep{zhou2022cocoop}. Encouragingly, using 1 shot for \method is already far better than using 16 shots for the compared baselines. \method with varying shots is also consistently better than the LLM-based but learning-free approach CuPL~\citep{pratt2023does}. Fig.~\ref{fig:scaling_property}(b) further shows that \method scales better with the number of used prompts than CuPL. Note when we use only 1 prompt generated by each LLM-prompt, CuPL is even worse than the CLIP baseline. Whereas \method performs much better by distilling task-related information from the limited number of prompts. Finally, Fig.~\ref{fig:scaling_property}(c) shows the benefits of \method over CuPL in terms of the scaling performance with LLM model size.

\begin{table*}[!t]
\caption{\textbf{Image captioning and VQA performance}. Note our \method is learned on COCO dataset, and we show both its zero-shot and finetuned results on the two tasks with different testing datasets.}
\label{tb:captioning_vqa}
% \vskip 0.05in
\centering
% \begin{center}
\resizebox{1.0\linewidth}{!}{
\begin{tabular}{lccc cccc cc cccc}
\toprule
 &\multicolumn{9}{c}{\textbf{Image Captioning}} & \multicolumn{4}{c}{\textbf{VQA2 K-shots}} \\
 \cmidrule(lr){2-10}
 \cmidrule(lr){11-14}
 & \multicolumn{3}{c}{COCO} & \multicolumn{4}{c}{NoCaps (CIDEr-D)} & \multicolumn{2}{c}{NoCaps (All)} & \multirow{2}{*}{0} & \multirow{2}{*}{1} & \multirow{2}{*}{2} & \multirow{2}{*}{4}  \\
\cmidrule(lr){2-4}
\cmidrule(lr){5-8}
\cmidrule(lr){9-10}
 &  CIDEr-D & CLIP-S & Ref-S & In & Out & Near & All & CLIP-S & Ref-S & & & & \\ \midrule
MAGMA~\citep{eichenberg2022magma} & 47.5 & 75.3 & 79.6 & 30.4 & 43.4 & 36.7 & 38.7 & 74.3 & 78.7 & 24.6& 39.3& 40.6& 41.5 \\
LiMBeR~\citep{merullo2023linearly} & 54.9 & 76.2 & 80.4 & 34.3 & 48.4 & 41.6 & 43.9 & 74.7 & 79.4 & 33.3& 39.9& 40.8& 40.3\\ \midrule
LiMBeR+\method (train/finetune) & \textbf{57.8} & \textbf{80.8} & \textbf{83.6} & \textbf{42.1} & \textbf{49.8} & \textbf{44.2} & \textbf{47.3} & \textbf{77.6} & \textbf{81.7}  & \textbf{36.5} & \textbf{42.7}& \textbf{44.2}& \textbf{45.9} \\
LiMBeR+\method (zero-shot) & - & - & - & 36.1 & 48.8 & 42.9 & 45.1 & 76.3 & 80.3 & 34.9 & 41.0& 42.3& 43.1 \\
\bottomrule
\end{tabular}
}
% \end{center}
\vskip -0.05in
\end{table*}

\subsection{Image Captioning \& VQA}
Table~\ref{tb:captioning_vqa} shows that \method can adapt to other vision-language tasks. Note \method is trained on COCO captions that describe complex scenes other than object-centric images. We once again find the benefits of distilling the text knowledge from image captions into \method. We observe consistent gains over the LiMBeR baseline, using either a trained or finetuned prompt generator (finetuned on NoCaps captioning and VQA2 tasks). More importantly, \method shows great generalization capability on both tasks. Its zero-shot performance is consistently better than that of LiMBeR and MAGMA, the latter of which finetunes both image and text networks. Fig.~\ref{fig:captioning_results} in Appendix~\ref{sec:captioning_results} exemplifies how \method can help on the captioning task, especially when the visual cues are ambiguous.

% stronger baselines: VLAP (https://arxiv.org/pdf/2404.09632), MAPL (https://arxiv.org/pdf/2210.07179) and DePALM (https://arxiv.org/pdf/2403.13499)

%data-deficient/low-data, few shot, domain gap/shift, OOD
% \method differs in downstream adaptation of the language priors through prompt learning.
\section{Conclusion}
\label{sec:Conclusion}

In this paper, we show the distillation of text-based knowledge into CLIP improves its downstream generalization. We propose a new prompt learning method where a prompt embedding \method is distilled from human- or LLM-generated natural language prompts. A prompt generator is trained to predict \method, which is shown to generalize to various vision-language tasks. We further demonstrate the benefits of \method for handling non-canonical examples as well as few-shot and OOD settings.

\paragraph{Limitations and future work.} Given a sufficiently large set of image prompts, it is preferable to aggregate them into more than one prompt embeddings to model text diversity. However, learning to predict such an embedding set is hard on data-deficient tasks, since it will not only increase the computation cost but also incur performance degradation. For future work we hope to address this limitation by scaling up data to learn a diversified universal prompt generator. Another plan is to go beyond CLIP and apply \method learning to more vision-language models (contrastive or generative).

\bibliography{myref}
\bibliographystyle{iclr2024_conference}

%%%%%%%%%%%%%%%%%%%%%%%%%%%%%%%%%%%%%%%%%%%%%%%%%%%%%%%%%%%%
\newpage
\appendix

\section{Alternative Methods of Generating Natural Language Prompts}

\textbf{Text retrieval for object-centric image classification.} It is possible to use the class name to retrieve class descriptions from WordNet~\citep{Miller95} or image-text datasets, just like in previous works~\citep{bujwid-sullivan-2021,shen2022k}. Unfortunately, no single dataset, including the large-scale LAION-5B~\citep{schuhmann2022laionb}, contains arbitrary classes specified in downstream tasks (\eg,~ fine-grained flower species). In other words, the retrieval-based approach can limit the possible classes that can be recognized. Furthermore, the image captions retrieved from image-text datasets can be noisy and irrelevant to the target class. One typical example is that the retrieved image captions are not really focused on the descriptive details of the target class instances. Instead, the captions can be depicting their relations with other objects in a multi-object image. Therefore, retrieved image captions may not be suited for object-centric classification.
This is evidenced by~\citet{pratt2023does} who explored similar retrieval ideas for ImageNet classification, where each class is guaranteed to have class descriptions from WordNet~\citep{Miller95} or Wikipedia articles (ImageNet-Wiki~\citep{bujwid-sullivan-2021}). The retrieved captions prove less effective than LLM generated customized prompts in terms of ImageNet Top-1 accuracy.

\textbf{Large Multimodal Models (LMMs) for object-centric classification and other vision-language tasks.}
One alternative way of acquiring natural language prompts is to query LMMs,~\eg,~GPT-4V~\citep{yang2023dawn}. For the task of object-centric classification, it is easy to imagine that LMMs can be more competent than pure text-based LLMs in generating helpful prompts, since LMMs have both vision and language supervision in their pretraining. For the same reason, vision-language tasks like VQA are likely to benefit more from LMM-generated image prompts than human-annotated image captions.
However, our goal here is to study the role of \emph{text-only} knowledge for downstream generalization. Hence LMMs with additional vision supervision fall outside of the scope of our study.

\section{More Ablations and Analyses}

\subsection{Input-Adapted Prompt Aggregator}
\label{sec:appendix_ablation_aggregator}

We rely on the input-adapted prompt aggregator to provide good textual supervision for prompt learning. The prompt aggregator is learned to aggregate all the reference prompts $\mP$ into an image-aligned prompt embedding $\vp^a$ with reduced noise and redundancy.

\begin{figure}[!t]
% \vskip -0.2in
\begin{center}
\centerline{\includegraphics[width=1.0\linewidth]{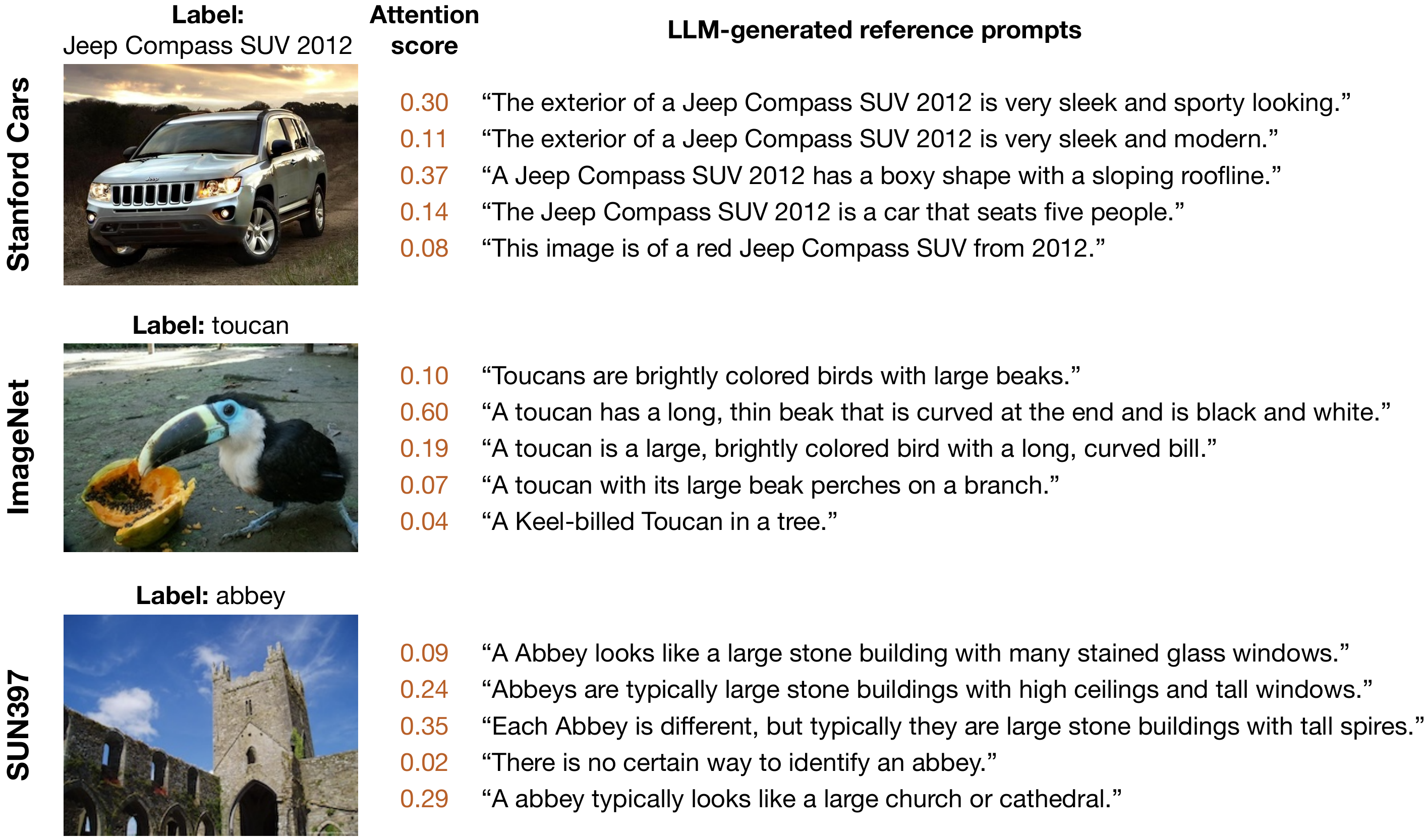}}
\caption{\textbf{Visualizing the attention score of each reference prompt during prompt aggregation}. Note the attention score is re-normalized among the illustrated prompt samples. We observe low attention scores for prompts that are redundant or noisy (irrelevant to input image),~\eg,~the 2nd and 5th prompt in the car image.}
\label{fig:prompt_score}
\end{center}
% \vskip -0.3in
\end{figure}

Fig.~\ref{fig:prompt_score} confirms that redundant and noisy (irrelevant) reference prompts are often suppressed with low \textbf{attention scores} during prompt aggregation.

Next, we \textbf{compare with three alternative methods for prompt aggregation}:
\vspace{-0.06in}
\begin{itemize}[leftmargin=20pt]
\setlength{\itemsep}{0pt}
\item Two baselines are learning-free, obtaining $\vp^a$ by either random sampling from $\mP$ or simple averaging (\ie,~$\bar{\vp}$).
\item  We further compare with a learning method using a different architecture other than the default attention network. Specifically, we start with the mean $\bar{\vp}$ and learn to transform it via simple MLP layers such that the transformed embedding is aligned with input image. Hence we have $\vp^a = \alpha f(\bar{\vp})+ (1-\alpha) \bar{\vp}$, where $f$ is a two-layer bottleneck MLP with ReLU nonlinearity, while $\alpha$ is a learnable parameter to weight the residual connection. Note such MLP-based architecture differs from attention mechanism in that the MLP-aggregated $\vp^a$ is not able to attend to individual prompts in $\mP$ for dynamic information fusion.
\end{itemize}

\begin{table*}[h]
\vskip -0.2in
\caption{\textbf{Ablation study on our prompt aggregator.} We experiment under the base-to-new class generalization setting for few-shot classification. CLIP performance (zero-shot) is listed as a baseline. H: Harmonic mean of base and new class accuracies.}
\label{tb:aggregator_ablation}
\begin{center}
\resizebox{0.78\linewidth}{!}{
\begin{tabular}{lccc}
\toprule
Avg across 11 datasets & Base & New & H \\
\midrule
Zero-shot CLIP~\citep{radford2021learning} & 69.34 & 74.22 & 71.70 \\ \midrule
Random sampling & 70.13\textsubscript{$\pm$0.93} & 74.61\textsubscript{$\pm$1.34} & 72.30\textsubscript{$\pm$1.15}\\
Mean $\bar{\vp}$ & 82.05\textsubscript{$\pm$0.13} & 75.64\textsubscript{$\pm$0.21} & 78.71\textsubscript{$\pm$0.15}\\
MLP-based aggregation & 84.39\textsubscript{$\pm$0.21} & 76.33\textsubscript{$\pm$0.28} & 80.16\textsubscript{$\pm$0.22}\\
Attention-based aggregation (default) & \textbf{84.72}\textsubscript{$\pm$0.18} & \textbf{77.54}\textsubscript{$\pm$0.29} & \textbf{80.97}\textsubscript{$\pm$0.19}\\
\bottomrule
\end{tabular}
}
\end{center}
% \vskip -0.1in
\end{table*}

Table~\ref{tb:aggregator_ablation} summarizes the comparison results on few-shot classification under the base-to-new class generalization setting. It is clear that random sampling is not a good choice, whose performance has large variance and is only marginally better than that of zero-shot CLIP. This is because the reference prompts are often noisy and not related to input image. Hence a randomly sampled prompt is likely a poor source of textual knowledge to distill from. When we use the mean prompt $\bar{\vp}$ as supervision, significant gains are observed due to reduced noise as well as enriched information. MLP-based aggregation leads to larger gains, since learning is introduced now to find a better and image-aligned supervisory signal. With the attention mechanism, we achieve the best results with dynamic prompt aggregation, which is our default approach.

\begin{table*}[t]
\vskip -0.3in
\caption{\textbf{Comparison with the $h(\vx)$-only baseline for few-shot classification.} We experiment under the base-to-new class generalization setting. H: Harmonic mean of base and new class accuracies.}
\label{tb:hx_only_classification}
\begin{center}
\resizebox{0.67\linewidth}{!}{
\begin{tabular}{clccc}
\toprule
& Avg across 11 datasets & Base & New & H \\
\midrule
\multirow{3}{*}{\rotatebox[origin=c]{90}{\shortstack[c]{LLM-\\based}}} & \method (default) & \textbf{84.72}\textsubscript{$\pm$0.18} & \textbf{77.54}\textsubscript{$\pm$0.29} & \textbf{80.97}\textsubscript{$\pm$0.19} \\ 
& (\method) $h(\vx)$-only & 84.01\textsubscript{$\pm$0.12} & 75.93\textsubscript{$\pm$0.16} & 79.77\textsubscript{$\pm$0.13}\\ 
& CuPL~\citep{pratt2023does} & 74.31\textsubscript{$\pm$0.00} & 75.25\textsubscript{$\pm$0.00} & 74.78\textsubscript{$\pm$0.00}\\ \midrule
\multirow{2}{*}{\rotatebox[origin=c]{90}{\shortstack[c]{w/o\\priors}}} & \method w/o $\mathcal{L}_{\text{distill}}$ & 79.47\textsubscript{$\pm$0.22} & 73.25\textsubscript{$\pm$0.34} & 76.23\textsubscript{$\pm$0.22}\\
& CoCoOp~\citep{zhou2022cocoop} & 80.47\textsubscript{$\pm$0.21} & 71.69\textsubscript{$\pm$0.37} & 75.83\textsubscript{$\pm$0.24}\\
\bottomrule
\end{tabular}
}
\end{center}
\vskip -0.1in
\end{table*}

\subsection{Simple Classification Framework Based on \method Only}
\label{sec:appendix_other_classification_framework}

In the main paper, we introduce our default classification framework in Section~\ref{sec:AAPE}. The classifier weights are built from the combination of \method $h(\vx)$ and a template embedding $\vw_{i}$ using a projection $g$. Here we compare with a simpler $h(\vx)$-only baseline for classification, without combining $\vw_{i}$ or using projection $g$. This baseline uses the standard text classifier weights $\vw_{i}$, while $h(\vx)$ acts as a proxy image query to be matched to $\vw_{i}$. Such setup is extremely similar to the image-to-text retrieval task. Evaluating the $h(\vx)$-only baseline is a more direct quantification of how well the text knowledge in \method, the image captioning embedding, can distinguish different classes.

Table~\ref{tb:hx_only_classification} shows the $h(\vx)$-only baseline achieves reasonable performance for both base and new classes, which indicates \method's good generalization on the classification task. When compared to CuPL that captures text knowledge by simply ensembling LLM-generated image prompts, the $h(\vx)$-only baseline is much more performant by learning an adaptive prompt aggregation. While the \method baseline w/o $\mathcal{L}_{\text{distill}}$ and CoCoOp both learn input-adapted prompts but without language supervision. We can observe evident benefits of our $h(\vx)$-only baseline over the language-free methods.

Lastly, our default \method-based classification framework consistently outperforms the $h(\vx)$-only baseline, only at a small overhead incurred by projection $g$. In the meantime, since the default classification framework simply combines and projects \method $h(\vx)$ and $\vw_{i}$, it enables easy interpretation of the roles of the two components for classification, see Fig.~\ref{fig:wi_AAPE}.

\section{Hyperparameter Sensitivity and Compute Cost}
\label{sec:appendix_hyperparameter_cost}

\paragraph{Sensitivity analysis of the distillation loss weight $\lambda$ in Eq.~(\ref{eq4}).} Table~\ref{tb:lambda_ablation} reports the results for the few-shot classification task. It is shown that \method performs robustly with overlapping confidence intervals when $\lambda \in [3,9]$. \method even outperforms the strong baseline OGEN (average H: 80.34) in this wide range of $\lambda$. We set $\lambda=5$ by default.

\begin{table*}[!t]
% \vskip -0.2in
\caption{\textbf{Sensitivity analysis of the distillation loss weight $\lambda$}. We report the few-shot classification results (base-to-new class generalization setting) averaged across 11 datasets in terms of H, the Harmonic mean of base and new class accuracies.}
\label{tb:lambda_ablation}
\begin{center}
\resizebox{0.9\linewidth}{!}{
\begin{tabular}{cc ccc ccc}
\toprule
$\lambda$ & 1 & 3 & 4 & 5 & 6 & 7 & 9 \\ \midrule
H & 79.14\textsubscript{$\pm$0.27} & 80.65\textsubscript{$\pm$0.23} & 80.84\textsubscript{$\pm$0.18} & \textbf{80.97}\textsubscript{$\pm$0.19} & 80.90\textsubscript{$\pm$0.14} & 80.93\textsubscript{$\pm$0.16} & 80.76\textsubscript{$\pm$0.21} \\
\bottomrule
\end{tabular}
}
\end{center}
\vskip -0.1in
\end{table*}

\begin{table*}[!t]
\caption{\textbf{Inference cost for few-shot classification and 3 other tasks beyond classification}. For few-shot classification, we report accuracy averaged over 11 datasets (base-to-new setting). We implement CoCoOp$\dagger$ to have a bigger prompt prediction network than CoCoOp, such that CoCoOp$\dagger$ has matching parameter count with \method.}
\label{tb:inference_cost}
\begin{center}
\resizebox{0.9\linewidth}{!}{
\begin{tabular}{lc ccc ccc}
\toprule
\multirow{2}{*}{Method} & \multirow{2}{*}{Comment} & \multicolumn{3}{c}{Compute cost} & \multicolumn{3}{c}{Few-shot accuracy} \\ \cmidrule(lr){3-5} \cmidrule(lr){6-8}
 & & \# params & GFLOP & FPS & Base & New & H \\ \midrule
CoOp~\citep{zhou2021coop} & \multirow{2}{*}{\shortstack[c]{\textbf{Non-adaptive}\\ \textbf{text prompts}}} & 2k & 162.5 & 1344 & 82.69 & 63.22 & 71.66 \\ 
OGEN~\citep{zang2024overcoming} & & 2k & 162.5 & 1351 & 84.17 & 76.86 & 80.34\\\midrule
MaPLe~\citep{khattak2023maple} & \multirow{2}{*}{\shortstack[c]{\textbf{Text+image}\\ \textbf{prompts}}} & 3.55M & 162.7 & 1365 & 82.28 & 75.14 & 78.55 \\
PromptSRC~\citep{khattak2023self} & & 46k & 162.8 & 1380 & 84.26 & 76.10 & 79.97 \\ \midrule
CoCoOp~\citep{zhou2022cocoop} &  & 35k & 162.5 & 15.08 & 80.47 & 71.69 & 75.83 \\ 
CoCoOp$\dagger$ &  & 84k & 162.6 & 14.67 & 81.05 & 70.12 & 75.19 \\ 
\rowcolor{gray!40}  & & & & & & & \\
\rowcolor{gray!40} \multirow{-2}{*}{\shortstack[l]{\method $h(\vx)$+$g$\\ for classification}} &  & \multirow{-2}{*}{82k} & \multirow{-2}{*}{162.6} & \multirow{-2}{*}{14.92} & \multirow{-2}{*}{\textbf{84.72}} & \multirow{-2}{*}{\textbf{77.54}} & \multirow{-2}{*}{\textbf{80.97}} \\
\rowcolor{black!30}  & & & & & & & \\
\rowcolor{black!30} \multirow{-2}{*}{\shortstack[l]{\method $h(\vx)$\\ beyond classification}} & \multirow{-5}{*}{\shortstack[c]{ \textbf{Input-adaptive}\\ \textbf{text prompts}}} & \multirow{-2}{*}{33k} & \multirow{-2}{*}{162.5} & \multirow{-2}{*}{15.06} & \multirow{-2}{*}{-} & \multirow{-2}{*}{-} & \multirow{-2}{*}{-} \\
% \rowcolor{gray!50} \method $h(\vx)$ (+$g$) & for classification & 83k & 162.6 & 14.92 & 84.72 & 77.54 & 80.97 \\
% \rowcolor{gray!50} \method $h(\vx)$ & beyond classification & 34k & 162.6 & 15.06 & - & - & - \\
\bottomrule
\end{tabular}
}
\end{center}
\vskip -0.1in
\end{table*}

\paragraph{Compute cost.} Table~\ref{tb:inference_cost} compares the inference cost for different tasks in terms of \# parameters, GFLOP and FPS. For the few-shot classification task, we compare \method with three types of prompt learning methods. CoOp and OGEN are the first type of methods that learn fixed text prompts. The efficiency benefits of these methods are evident: the number of learned parameters is small (2k), and high inference speed (FPS) can be achieved without requiring a forward pass to predict adaptive prompts for every input image. MaPLe and PromptSRC belong to the multimodal prompting methods that learn prompts for both text and image. These methods have comparable GFLOP and FPS with fixed prompt learners but have much more parameters to learn, thus risk generalization with sub-optimal accuracy for new classes.

CoCoOp and our \method both learn input-adaptive text prompts, with reasonable parameter count and GFLOP. However, they suffer from low FPS because of the input-conditional prompt prediction. This low speed also translates to the learning stage. The training time (min) for \method and CoCoOp are 41.92 and 39.53 respectively, in comparison to 10.08 of CoOp.
Despite the equally low time efficiency, \method outperforms CoCoOp drastically in accuracy, and scales much better with model size than CoCoOp-style methods. To show this, we implement a CoCoOp$\dagger$ baseline that has a similar parameter count with \method. As expected, CoCoOp$\dagger$ has lower speed than CoCoOp. CoCoOp$\dagger$ is also found to have lower new class accuracy,~\ie,~worse generalization.

When we turn our attention to complex vision-language tasks beyond classification, \method shines in both efficiency and performance. Note in the tasks of text retrieval, image captioning and VQA, \method is used as a standalone image captioning embedding (without projection $g$) to provide rich language priors. As shown in Table~\ref{tb:inference_cost} (last row), this setup involves fewer parameters than the classification setting (33k vs. 82k), and leads to slightly higher inference speed. But on the vision-language tasks, \method not only achieves SOTA performance (Table~\ref{tb:captioning_vqa}) but also has higher efficiency than prior works,~\eg, about 2.8/1.2 times faster than MAGMA~\citep{eichenberg2022magma} for training/inference.

In summary, \method is designed to be a universal text embedding directly applicable to various vision-language tasks. This is not possible with most prompting methods designed for classification, only that \method's generality sacrifices the efficiency in the classification task. We leave as future work to speed up AAPE inference in the classification setting,~\eg,~via pruning or distillation techniques to simplify the forward pass of prompt prediction.

\begin{table*}[!t]
\caption{\textbf{Few-shot classification in the domain generalization setting.} Note our AAPE follows CuPL to query an LLM to obtain natural language prompts, but further learns from those prompts.}
\label{tb:domain_generalization}
% \vskip -0.1in
\begin{center}
\resizebox{1.0\linewidth}{!}{
\begin{tabular}{clccccc}
\toprule
& & \multicolumn{1}{c}{Source} & \multicolumn{4}{c}{Target} \\
\cmidrule(lr){3-3}
\cmidrule(lr){4-7}
& & ImageNet & ImageNetV2 & ImageNet-Sketch & ImageNet-A & ImageNet-R \\
\midrule
Zero-shot & CLIP~\citep{radford2021learning} & 66.73 & 60.83 & 46.15 & 47.77 & 73.96\\ \midrule
\multirow{8}{*}{\rotatebox[origin=c]{90}{\shortstack[c]{Prompt learning w/o\\language priors}}} & MaPLe~\citep{khattak2023maple} & 70.72 & 64.07 & 49.15 & 50.90 & 76.98\\
& CoCoOp~\citep{zhou2022cocoop} & 71.02 & 64.07 & 48.75 & 50.63 & 76.18\\
& PromptSRC~\citep{khattak2023self} & 71.27 & 64.35 & 49.55 & 50.90 & \textbf{77.80}\\
& CoOp~\citep{zhou2021coop} & 71.51 & 64.20 & 47.99 & 49.71 & 75.21\\
& CLIPood~\citep{shu2023clipood}  & 71.60 & 64.90 & 49.30 & 50.40 & 77.20\\
& RPO~\citep{lee2023rpo}  & 71.67 & 65.13 & 49.27 & 50.13 & 76.57\\
& UPT~\citep{zang2022unified}  & 72.63 & 64.35 & 48.66 & 50.66 & 76.24\\
& TaskRes~\citep{yu2023task}  & 73.07 & 65.30 & 49.13 & 50.37 & 77.70\\ \midrule
\multirow{3}{*}{\rotatebox[origin=c]{90}{\shortstack[c]{Basic\\prompts}}} & LASP~\citep{bulat2023lasp} & 71.10 & 63.96 & 49.01 & 50.70 & 77.07\\
 & KgCoOp~\citep{yao2023visual} & 71.20 & 64.10 & 48.97 & 50.69 & 76.70\\
 & ProGrad~\citep{zhuBeier2023} & 72.24 & 64.73 & 47.61 & 49.39 & 74.58\\ \midrule
\multirow{2}{*}{\rotatebox[origin=c]{90}{\shortstack[c]{LLM-\\based}}} & CuPL~\citep{pratt2023does} & 68.86 & 63.14 & 47.85 & 48.63 & 75.11\\
& \method & \textbf{73.56}\textsubscript{$\pm$0.12} & \textbf{65.97}\textsubscript{$\pm$0.18} & \textbf{50.12}\textsubscript{$\pm$0.23} & \textbf{51.62}\textsubscript{$\pm$0.22} & 77.52\textsubscript{$\pm$0.14}\\ [0.5ex]
\bottomrule
\end{tabular}
}
\end{center}
\vskip -0.1in
\end{table*}

\section{Results of Few-Shot Classification under Domain Generalization}
\label{sec:domain_generalization}
Table~\ref{tb:domain_generalization} shows our approach is robust to the different types of domain shifts on 4 ImageNet variants. Overall, our \method outperforms prior works on all but the ImageNet-R dataset (where \method is still competitive with the state-of-the-art PromptSRC method). \method outperforms most prompt learners that do not leverage any language priors, sometimes by a large margin. When compared to the prompt learners using basic prompt templates, \method shows notable gains thanks to the rich knowledge contained in LLMs. The zero-shot CuPL method is based on LLM too, but lags far behind due to the lack of learning components for downstream adaptation.

\section{Comparison with Recent Prompt Learners ProText and ArGue-N}
\label{sec:ProText_ArGue}

Table~\ref{tb:ProText_ArGue} compares \method with two recent prompt learning methods on the few-shot classification task. The compared methods ProText and ArGue-N are closely related to \method since the former similarly learn text prompts from LLM-generated image prompts or visual attributes. However, both ProText and ArGue-N learn fixed text prompts, and are not adapted to input image. By contrast, \method is input-adaptive during both prompt aggregation and prompt prediction -- recall that our prompt aggregation is aligned with the input image, and \method prediction is image-conditional.

We conjecture that our input-adaptive framework makes better use of LLM's textual knowledge. The resulting image-aligned \method also promotes image-text alignment, giving rise to improved optimization during prompt learning. Table~\ref{tb:ProText_ArGue} confirms this with better performance of \method on seen class data from the base split or source dataset. Input-adaptive \method improves generalization too, achieving comparable or stronger generalization performance in two generalization settings than the fixed prompt learners. More importantly, \method can be used as a standalone captioning vector in various vision-language tasks beyond classification. This is not possible with ProText or ArGue-N.

\begin{table*}[t]
% \vskip -0.3in
\caption{\textbf{Comparison with ProText and ArGue-N} for few-shot classification in both settings of base-to-new class generalization and domain generalization.}
\label{tb:ProText_ArGue}
\begin{center}
\resizebox{0.9\linewidth}{!}{
\begin{tabular}{lcccccccc}
\toprule
& \multicolumn{3}{c}{\textbf{Base-to-New Generalization}} & \multicolumn{5}{c}{\textbf{Domain Generalization}} \\
\cmidrule(lr){2-4} \cmidrule(lr){5-9}
& \multicolumn{3}{c}{Avg across 11 datasets} & \multicolumn{1}{c}{Source} & \multicolumn{4}{c}{Target} \\
\cmidrule(lr){2-4} \cmidrule(lr){5-5} \cmidrule(lr){6-9}
& Base & New & H & ImageNet & -V2 & -Sketch & -A & -R \\
\midrule
ProText~\citep{Khattak2024ProText} & 72.95 & 76.98 & 74.91 & 70.22 & 63.54 & 49.45 & 51.47 & 77.35 \\
ArGue-N~\citep{10657279} & 83.77 & \textbf{78.74} & \textbf{81.18} & 71.84 & 65.02 & 49.25 & 51.47 & 76.96 \\
\method (ours) & \textbf{84.72} & 77.54 & 80.97 & \textbf{73.56} & \textbf{65.97} & \textbf{50.12} & \textbf{51.62} & \textbf{77.52}\\
\bottomrule
\end{tabular}
}
\end{center}
% \vskip -0.1in
\end{table*}

\section{Additional Image Captioning Results}
\label{sec:captioning_results}

\begin{figure}[!t]
% \vskip -0.3in
\begin{center}
\centerline{\includegraphics[width=0.93\linewidth]{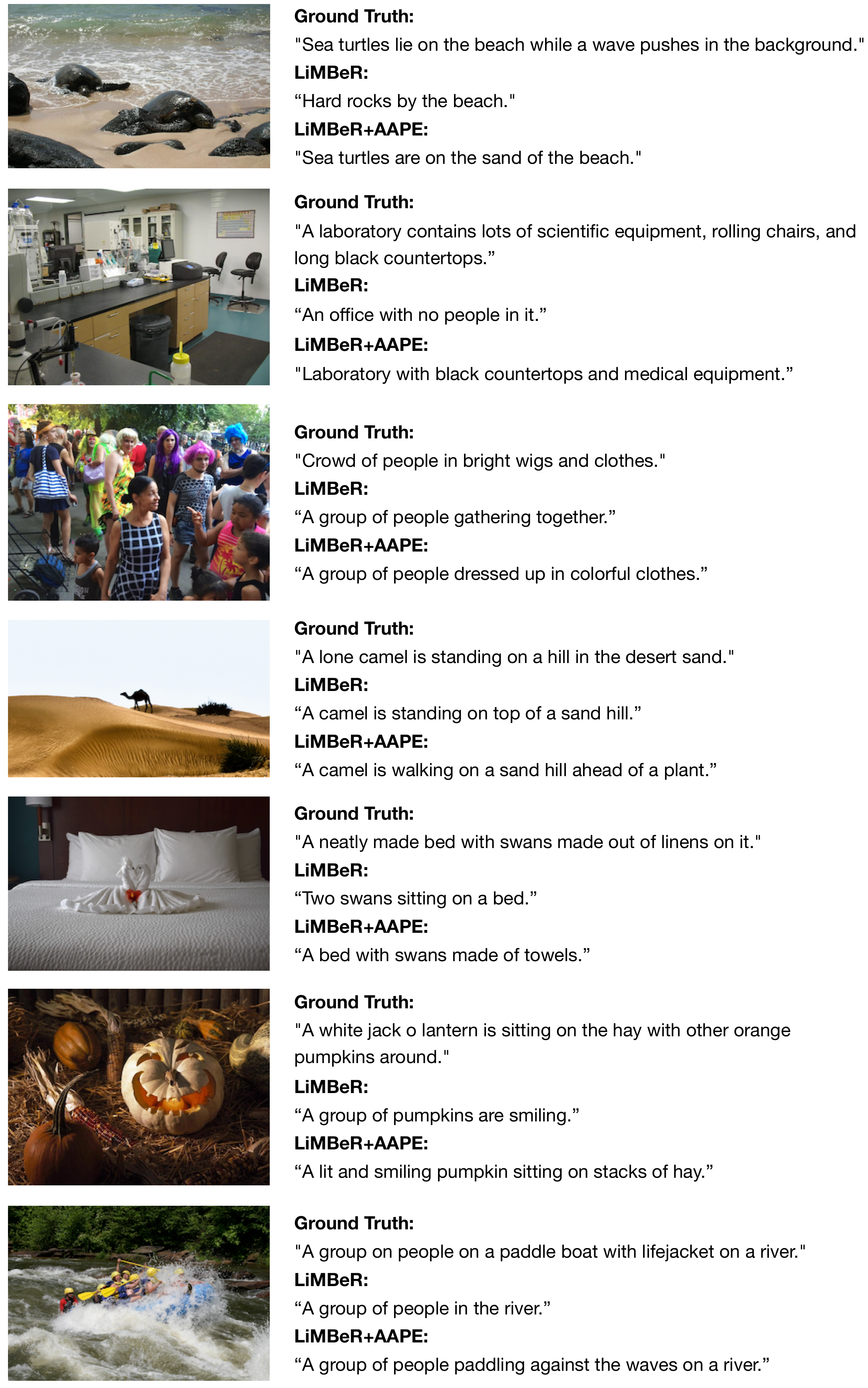}}
\caption{\textbf{Captioning results on NoCaps dataset}: LiMBeR vs. LiMBeR+AAPE (zero-shot).}
\label{fig:captioning_results}
\end{center}
% \vskip -0.2in
\end{figure}

Fig.~\ref{fig:captioning_results} shows example captioning results on the NoCaps dataset. We compare LiMBeR~\citep{merullo2023linearly} with and without our \method learned on COCO and transferred zero-shot to NoCaps. Thanks to the textual knowledge encoded in \method, it often augments the image features to generate more descriptive captions even when the visual cues are ambiguous (\eg,~in the first image, \method identifies the sea turtles which can be easily confused with rocks).

\begin{figure}[!t]
\begin{center}
\centerline{\includegraphics[width=1.0\linewidth]{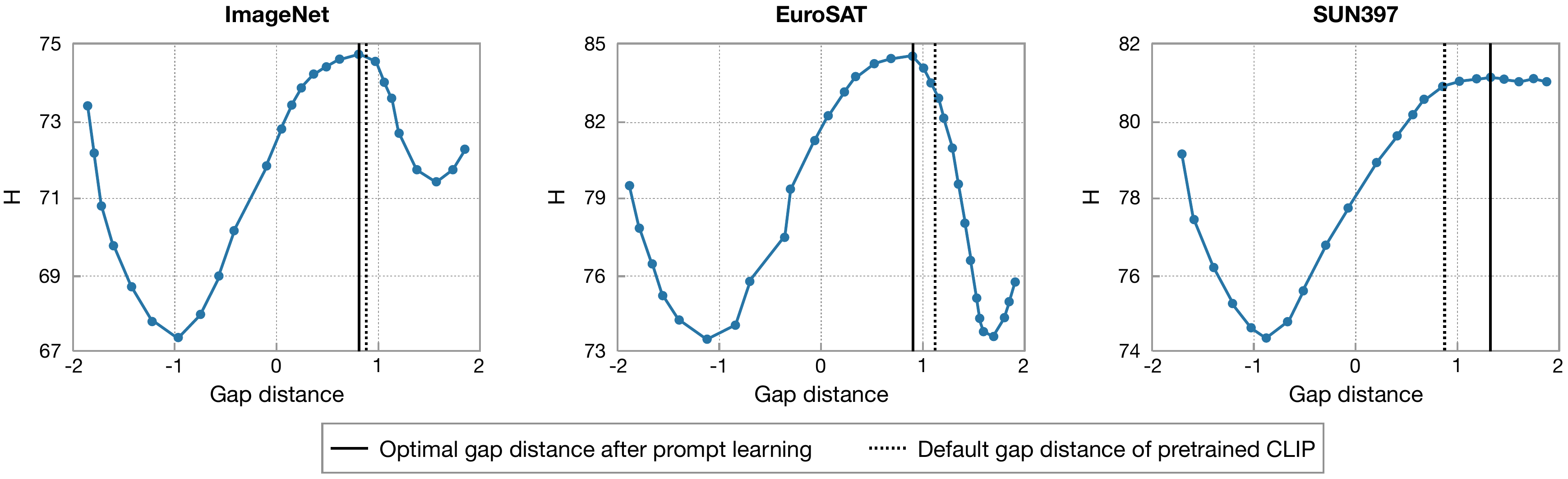}}
\caption{\textbf{Optimal gap distance obtained by \method learning} on the few-shot classification task (base-to-new class generalization setting). Y axis indicates the Harmonic mean (H) of base and new class accuracies on each dataset. X axis indicates the gap distance varied by shifting the image and text embeddings following~\citep{ModalityGap}. We see the optimal gap distance can be increased (on SUN397) or decreased (on ImageNet and EuroSAT) over the default gap obtained from pretraining.}
\label{fig:modality_gap}
\end{center}
\vskip -0.2in
\end{figure}

\section{Discussion on the Image-Text Modality Gap for \method Learning}

In this section, we examine the image-text modality gap, a concept introduced in~\citep{ModalityGap}. We aim to gain insights behind the strong downstream performance and generalization of \method-based prompt learning. Is that because \method learning can mitigate modality gap?

To answer this question, we first note that \method is predicted from input image features $\vx$ via a conditional prompt generator $h$. Such $h$ can be viewed as an image-to-text mapping function, which should at least maintain (if not improve) feature alignment between the image and text modalities.
For empirical evidence, we measure the average cosine feature similarity for the image-prompt pairs on ImageNet. We find \method scores 0.91, slightly higher than that of pretrained CLIP (0.89).
This fact makes it seem like \method should be able to reduce modality gap. Here we perform a systematic analysis following~\citep{ModalityGap}, which defines modality gap as the \textit{Euclidean distance between the centers of image and text features}.

Fig.~\ref{fig:modality_gap} shows the analysis results for few-shot classification on three example datasets. As expected, after \method-based prompt learning, the optimal gap distance that attains maximum accuracy deviates from the default gap distance of pretraind CLIP. We see the optimal gap is increased on SUN397 dataset, while decreased on ImageNet and EuroSAT datasets. In fact, the optimal gap is reduced on 7 out of a total of 11 datasets and moderately increased on the remaining 4. On the other hand, \method consistently improves classification accuracy on each dataset, see Table~\ref{tb:base2new} and Fig.~\ref{fig:base2new_gain}. This suggests that \textbf{modality gap is not highly correlated with downstream generalization}, which is in line with one of the main arguments in~\citep{ModalityGap} that \textbf{good generalization does not necessarily need a reduced modality gap}.

\textbf{Then why does \method generalize when the modality gap is not reduced?} We hypothesize that our multi-task learning reshapes the loss landscape in a way that encourages generalizable solutions. Specifically, we optimize two loss functions for \method: the task loss that (over)fits the seen class data, and the distillation loss that moves \method closer to some aggregation of external text knowledge. We find the distillation loss often promotes generalization (Fig.~\ref{fig:base2new_gain}) and avoids overfitting caused by the task loss. Hence the two losses could move \method in opposite directions, modifying the modality gap differently on the loss landscape. As shown in Fig.~\ref{fig:modality_gap}, the gap change (increase or decrease) is highly dependent on the image-text distribution on the considered dataset.

Note the above optimization perspective is not limited to classification. We can use our hypothesis to similarly explain \method's good generalization in complex vision-language tasks like VQA where a multi-task loss is used (distillation + task loss). We leave as future work to study 1) how multi-task learning affects modality gap dynamically and 2) the relationship between modality gap and downstream generalization.

\section{Broader Impact}
\label{sec:broader_impact}

The main contribution of this work is the use of text-based knowledge to improve the downstream generalization of CLIP. The textual knowledge is distilled from either human-annotated image captions or LLM-generated natural language prompts. Such knowledge significantly improves CLIP's downstream performance but carries potential societal impacts. Specifically, when the image captions/prompts reflect (unintentional) biases, our distillation and learning methods could inherit or amplify these biases in the learned feature embeddings. This would potentially lead to perpetual unfair or discriminative outcomes in a variety of vision-language tasks and more critical applications such as AI-driven planning and decision-making.

%%%%%%%%%%%%%%%%%%%%%%%%%%%%%%%%%%%%%%%%%%%%%%%%%%%%%%%%%%%%

\newpage
\section*{NeurIPS Paper Checklist}

\begin{enumerate}

\item {\bf Claims}
    \item[] Question: Do the main claims made in the abstract and introduction accurately reflect the paper's contributions and scope?
    \item[] Answer: \answerYes{} % Replace by \answerYes{}, \answerNo{}, or \answerNA{}.
    \item[] Justification: The main claims are stated in both Abstract and Introduction sections, and each claim is supported by experimental results. We further enumerate our contributions in the Introduction section.

\item {\bf Limitations}
    \item[] Question: Does the paper discuss the limitations of the work performed by the authors?
    \item[] Answer: \answerYes{} % Replace by \answerYes{}, \answerNo{}, or \answerNA{}.
    \item[] Justification: We discuss the limitations and future work in the Conclusion section~\ref{sec:Conclusion}.

\item {\bf Theory Assumptions and Proofs}
    \item[] Question: For each theoretical result, does the paper provide the full set of assumptions and a complete (and correct) proof?
    \item[] Answer: \answerNA{} % Replace by \answerYes{}, \answerNo{}, or \answerNA{}.
    \item[] Justification: No theoretical results are included in the paper.

    \item {\bf Experimental Result Reproducibility}
    \item[] Question: Does the paper fully disclose all the information needed to reproduce the main experimental results of the paper to the extent that it affects the main claims and/or conclusions of the paper (regardless of whether the code and data are provided or not)?
    \item[] Answer: \answerYes{} % Replace by \answerYes{}, \answerNo{}, or \answerNA{}.
    \item[] Justification: Experimental setup and dataset details are included in Section~\ref{sec:setup_datasets}.

\item {\bf Open access to data and code}
    \item[] Question: Does the paper provide open access to the data and code, with sufficient instructions to faithfully reproduce the main experimental results, as described in supplemental material?
    \item[] Answer: \answerNo{} % Replace by \answerYes{}, \answerNo{}, or \answerNA{}.
    \item[] Justification: Section~\ref{sec:setup_datasets} describes the experimental setup, dataset and implementation details to run and reproduce experiments.

\item {\bf Experimental Setting/Details}
    \item[] Question: Does the paper specify all the training and test details (e.g., data splits, hyperparameters, how they were chosen, type of optimizer, etc.) necessary to understand the results?
    \item[] Answer: \answerYes{} % Replace by \answerYes{}, \answerNo{}, or \answerNA{}.
    \item[] Justification: Section~\ref{sec:setup_datasets} describes all the experimental setup, dataset and training/testing details.

\item {\bf Experiment Statistical Significance}
    \item[] Question: Does the paper report error bars suitably and correctly defined or other appropriate information about the statistical significance of the experiments?
    \item[] Answer: \answerYes{} % Replace by \answerYes{}, \answerNo{}, or \answerNA{}.
    \item[] Justification: We report the average result and the standard deviation of three runs with random seeds for our experiments/ablations on image classification and text retrieval. Please refer to Table~\ref{tb:text_retrieval},~\ref{tb:base2new},~\ref{tb:aggregator_ablation},~\ref{tb:hx_only_classification},~\ref{tb:lambda_ablation} and~\ref{tb:domain_generalization}.

\item {\bf Experiments Compute Resources}
    \item[] Question: For each experiment, does the paper provide sufficient information on the computer resources (type of compute workers, memory, time of execution) needed to reproduce the experiments?
    \item[] Answer: \answerYes{} % Replace by \answerYes{}, \answerNo{}, or \answerNA{}.
    \item[] Justification: This information is included in Section~\ref{sec:setup_datasets} and Appendix~\ref{sec:appendix_hyperparameter_cost}.
    
\item {\bf Code Of Ethics}
    \item[] Question: Does the research conducted in the paper conform, in every respect, with the NeurIPS Code of Ethics \url{https://neurips.cc/public/EthicsGuidelines}?
    \item[] Answer: \answerYes{} % Replace by \answerYes{}, \answerNo{}, or \answerNA{}.
    \item[] Justification: We confirm that the work presented in this paper is performed in a manner consistent with NeurIPS Ethics Guidelines.

\item {\bf Broader Impacts}
    \item[] Question: Does the paper discuss both potential positive societal impacts and negative societal impacts of the work performed?
    \item[] Answer: \answerYes{} % Replace by \answerYes{}, \answerNo{}, or \answerNA{}.
    \item[] Justification: We discuss the broader impact of our work in Appendix~\ref{sec:broader_impact}.
    
\item {\bf Safeguards}
    \item[] Question: Does the paper describe safeguards that have been put in place for responsible release of data or models that have a high risk for misuse (e.g., pretrained language models, image generators, or scraped datasets)?
    \item[] Answer: \answerNA{} % Replace by \answerYes{}, \answerNo{}, or \answerNA{}.
    \item[] Justification: Our work presents a finetuning algorithm of CLIP models so this question is not applicable.

\item {\bf Licenses for existing assets}
    \item[] Question: Are the creators or original owners of assets (e.g., code, data, models), used in the paper, properly credited and are the license and terms of use explicitly mentioned and properly respected?
    \item[] Answer: \answerYes{} % Replace by \answerYes{}, \answerNo{}, or \answerNA{}.
    \item[] Justification: We include this information in experimental details in Section~\ref{sec:setup_datasets}.

\item {\bf New Assets}
    \item[] Question: Are new assets introduced in the paper well documented and is the documentation provided alongside the assets?
    \item[] Answer: \answerNA{} % Replace by \answerYes{}, \answerNo{}, or \answerNA{}.
    \item[] Justification: This paper does not release new assets.

\item {\bf Crowdsourcing and Research with Human Subjects}
    \item[] Question: For crowdsourcing experiments and research with human subjects, does the paper include the full text of instructions given to participants and screenshots, if applicable, as well as details about compensation (if any)? 
    \item[] Answer: \answerNA{} % Replace by \answerYes{}, \answerNo{}, or \answerNA{}.
    \item[] Justification: This paper does not involve crowdsourcing nor research with human subjects.

\item {\bf Institutional Review Board (IRB) Approvals or Equivalent for Research with Human Subjects}
    \item[] Question: Does the paper describe potential risks incurred by study participants, whether such risks were disclosed to the subjects, and whether Institutional Review Board (IRB) approvals (or an equivalent approval/review based on the requirements of your country or institution) were obtained?
    \item[] Answer: \answerNA{} % Replace by \answerYes{}, \answerNo{}, or \answerNA{}.
    \item[] Justification: This paper does not involve crowdsourcing nor research with human subjects.

\end{enumerate}

\end{document}